%% file: main.tex
\pdfoutput=1
\documentclass[conference]{IEEEtran}
\IEEEoverridecommandlockouts
\usepackage[numbers]{natbib}
\usepackage{amsmath,amssymb,amsfonts}
\usepackage{algorithmic}
\usepackage{graphicx}
\usepackage{textcomp}
\usepackage{xcolor}
\usepackage{hyperref}
\usepackage{physics}
\usepackage[]{algorithm2e}

\def\BibTeX{{\rm B\kern-.05em{\sc i\kern-.025em b}\kern-.08em
    T\kern-.1667em\lower.7ex\hbox{E}\kern-.125emX}}

\begin{document}
\title{Deep Fusion of Lead-lag Graphs:\\Application to Cryptocurrencies}

\author{\IEEEauthorblockN{ Hugo Schnoering}
\IEEEauthorblockA{\textit{Napoleon Group} \\
hugo.schnoering@napoleon-group.com}
\and
\IEEEauthorblockN{Hugo Inzirillo}
\IEEEauthorblockA{\textit{CREST Institut Polytechnique de Paris} \\
\textit{Napoleon Group}\\
hugo.inzirillo@napoleon-group.com}
}
\maketitle

\begin{abstract}
    The study of time series has motivated many researchers, particularly on the area of multivariate-analysis. The study of co-movements and dependency between random variables leads us to develop metrics to describe existing connection between assets. The most commonly used are correlation and causality. Despite the growing literature, some connections remained still undetected. The objective of this paper is to propose a new representation learning algorithm capable to integrate synchronous and asynchronous relationships. 
\end{abstract}

\section{Introduction}

\input{sections/intro}

\section{Methods}

\textbf{Notations}:
\begin{itemize}
    \item $n$  the number of assets
    \item $p + 1$ the number of dates
    \item $\{t_\tau\}_{\tau=0}^{p} \triangleq \{t_0, t_1, t_2, ..., t_p\}$ the dates arranged in the chronological order
    \item $d$ the common time span between two consecutive dates
    \item $P_j(t_i)$ the price of asset $j$ at time $t_i$
     \item $r_j(t_i)$ the log-return of asset $j$ between $t_{i-1}$ and $t_i$
    \item $\mathbb{E}(X)$ is the expected value of the random variable $X$
    \item $\bra{u}\ket{v}$ is the dot product between vectors $u$ and $v$
    \item $|| u ||$ is the Frobenius norm of vector $ u $
    \item Let $\mathbf{M}$ a matrix. $\mathbf{M}_p$ is the $p$-th row of $\mathbf{M}$, $\mathbf{M}_{[p, q]}$ is the submatrix of $\mathbf{M}$ composed of rows $p, p+1, ..., q-1, q$, and $\mathbf{M}_{-, p}$ is the $p$-th column of $\mathbf{M}$. 
\end{itemize}

\subsection{Lead-lag Graph}
\input{sections/lead_lag_networks}

\subsection{DeepNF}
\input{sections/deepNF}

\subsection{Deep Fusion of Lead-lag Graphs}

\input{sections/deep_fusion_lead_lag_networks}

\section{Materials and Results}

\input{sections/material}

\section{Discussion}

\input{sections/discussion}

\section{Conclusion}
\input{conclusion}

\bibliographystyle{IEEEtranN}
\bibliography{bib}
\nocite{*}

\section*{Appendix}

\input{sections/appendix}

\end{document}

%% file: sections/intro.tex





Recently, cryptocurrencies have become a center of interest for key players in financial markets and financial institutions. Despite this strong interest in cryptocurrencies,  the youth of these new digital assets makes them difficult to model. Connections between pairs are not well established. Several papers point out common behaviour patterns between quotes of several cryptocurrencies \citep{gkillas2018extreme, stosic2018collective}.
Many efforts have already been devoted to understanding correlations in financial markets and joint dynamics.  A large number of research papers are concentrated on Pearson’s correlation coefficient to detect synchronous relations on equity returns \citep{buccheri2013evolution, tumminello2010correlation} and cryptocurrency returns \citep{lahajnar2020correlation}. There are however two limits to this approach: 
\begin{itemize}
    \item it is only sensitive to linear dependencies,
    \item temporality is broken and it is therefore not sensitive to assets driving each other asynchronously, i.e. to lead-lag relationships. 
\end{itemize}

Statistical analyses using only correlation might not detect any connection between assets. Complexity of financial markets suggests that relationships between assets are not bound to be linear. Several papers already support this hypothesis \citep{brock1991nonlinear, sornette2002nonlinear}. In order to detect non linear relations, more general measures may be used: e.g. the Spearman's rank correlation or the Kendall’s rank correlation. Concerning the lead-lag relationships between assets, they cannot exist in the framework introduced by the standard financial theory. In particular, the Efficient Market Hypothesis (EMH) of Fama \cite{fama1960efficient} states that prices contain all available information. Consequently, it is not possible to predict future returns from the past. It has been therefore demonstrated in several works that this hypothesis is not verified in practice \citep{lu2012tests, nagayasu2003efficiency}. Interest of some researchers has thus shifted from the analysis of synchronous relationships to the analysis of asynchronous relationships.

\citeauthor{curme2015emergence} \cite{curme2015emergence} study lagged linear relationships between $i$ and $j$ by computing the correlation coefficient between the returns of $i$ and the forward-shifted returns of $j$. \citeauthor{fiedor2014information} \cite{fiedor2014information} undertakes the same methodology of \citeauthor{curme2015emergence} and replaces the correlation coefficient with the \emph{mutual information} in order to detect non-linear relations between the returns of $i$ and the lagged returns of $j$. Given a lag $T$ and a set of assets, \citeauthor{fiedor2014information} even constructs a \emph{lead-lag graph} in which assets are nodes, and edges represent validated relationships between assets. Such a graph accounts for relations that may exist between assets when information takes $T$ to be propagated. \citeauthor{fiedor2014information} also introduces a statistical test to determine whether a found relationship is statistically significant or not. This is done by computing a p-value for each lagged relationship, and then setting a threshold on this p-values.

Empirical results in \cite{fiedor2014information} suggest that the sampling frequency of prices denoted by $d$ plays an important role: relationships between assets are not consistent with the sampling frequency. Our first contribution is to investigate whether statistically significant lagged relationships exist in cryptocurrency markets for sampling periods equal to 1 minute and 5 minutes.

 An asset $x$ can be described in different manners: its advisory board, its supply network, its last returns, etc. Finding an embedding / representation of $x$ living in a Euclidean space can be useful because it allows to do basic arithmetic with $x$. Lots of distance can thus be used to compare two assets, opening up the possibility for clustering.   Given a lead-lag graph $\mathcal{G}$ characterized by $(T, d)$, representations of its nodes / assets can be derived from the graph $\mathcal{G}$ with various approaches such as \emph{Random Walk} \cite{spitzer2013principles} or \emph{Node2vec} \cite{grover2016node2vec}. Representation of an asset should however not be limited to a lag $T$ and a frequency $d$, but should take into account the different relationships that may exist between the assets. Given a set of lead-lag graphs obtained with the methodology of \citeauthor{fiedor2014information} \cite{fiedor2014information}, the main contribution of this article is to learn a fused representation of the assets from several lead-lag graphs.

Several works are dedicated to "fusing graphs" \citep{khan2019multi, zhang2020multiplex, shi2018mvn2vec}. In this paper, we use the model \emph{deepNF} introduced by \citeauthor{gligorijevic2018deepnf} \cite{gligorijevic2018deepnf}. Given a set of graphs sharing the same set of nodes, \emph{deepNF} learns in an unsupervised manner node where the representation accounts for information taken from the different input graphs. 

Finally, relations between assets may vary over time. This is especially true in our case, because cryptocurrency markets are in the early adoption phase, and new projects emerge every day. In our approach, asset representations are regularly updated by taking into account only the near past. A cosine similarity can then be used to monitor the evolution of the similarity between a pair of assets over time. From this similarity measure, we can also derive a similarity matrix which could be used for several purposes.

%% file: sections/lead_lag_networks.tex
Here, we present the methodology of \cite{fiedor2014information}  to identify non linear dependencies between financial instruments, which itself extends the methodology of  \cite{curme2015emergence}. The first step is the computation of the matrix $\mathbf{R}$ of log-returns for the different assets:

\begin{equation}
    \mathbf{R}_{i,j} = r_j(t_i) = \log(P_j(t_i))-\log(P_j(t_{i-1}))
    \label{eq:return_matrix}
\end{equation}

Let $T \in \mathbb{N}$ be the lag that will be used for forward shifting. $\mathbf{R}$ is filtered into two matrices, $\mathbf{A}^{(T)}$ and $\mathbf{B}^{(T)}$, in which the last $T$ returns are excluded from $\mathbf{A}^{(T)}$ and the $T$ first returns are excluded from $\mathbf{B}^{(T)}$. From these matrices the matrix $\mathbf{C}$ is constructed using \textit{Mutual Information} of columns of $\mathbf{A}^{(T)}$ and $\mathbf{B}^{(T)}$: 

\begin{equation}
    C_{m,n} := I_S(\mathbf{A}_{-, m}^{(T)}, \mathbf{B}_{-, n}^{(T)})
\end{equation}

where $I_S(\cdot, \cdot)$ is the \textit{Mutual Information} operator between two random variables. \textit{Mutual Information} for two discrete random variables $X$ and $Y$ is defined by equation \ref{eq:mutual_information}. 

\begin{equation}
\begin{split}
        I_S(X, Y) &  = D_{KL}\Big(p(x, y) || p(x)p(y)\Big)\\ 
        & = \mathbb{E}_{p(x,y)} \log \left( \frac{p(x, y)}{p(x)p(y)} \right)\\
        & = \sum_{y \in \mathcal{Y}} \sum_{x \in \mathcal{X}} p(x, y) \log \left( \frac{p(x, y)}{p(x)p(y)} \right), 
\end{split}
    \label{eq:mutual_information}
\end{equation}

where $p(x, y)$ is the joint probability distribution  of $(X, Y)$,  $p(x)$ and $p(y)$ are the marginal probability distributions, and $\mathcal{X} $ and $\mathcal{Y}$ are the sets of values that $X$ and $Y$ can take respectively. For continuous variables the definition remains valid by using integrals and probability density functions. It is straightforward to demonstrate that the mutual information between two independent variables is equal to 0. In addition, the mutual information is non negative. For this reason, if there is no statistical relation between the lagged returns $A_m$ and the returns $B_n$, we expect $C_{m, n}$ to be approximately equal to 0. \\

Even if log-returns are continuous variables, they are discretized for estimating the mutual information. Following \cite{fiedor2014information}, the mutual information is estimated with the plug-in estimator, which is the mutual information between the discretized empirical distributions. Once the coefficient $C_{i,j}= I_S(A_i, B_j)$ has been estimated by using the plug-in estimator $\hat{C}_{i,j}$, it is important to determine whether this coefficient is significant or not. As in \cite{fiedor2014information}, we use an approximation of the mutual information between two random variables by a gamma distribution in order to build a statistical test of significance. It has been shown in \cite{goebel2005approximation} that the mutual information between independent discrete random variables $X$ and $Y$ when estimated from relative frequencies follows a very good approximation of Gamma distribution $\mathcal{G}$ with parameters $$\alpha = (|\mathcal{X}|-1)(|\mathcal{Y}| - 1)/2 \quad \text{and} \quad \beta = 1/(N \log(2))$$ where $N$ is the sample size and $|\mathcal{X}|$ and $|\mathcal{Y}|$ denote the numbers of realizations of $X$ and $Y$ respectively. Given a threshold $p$, the statistical test thus checks the condition \ref{eq:signi_condition}. 

\begin{equation}
    \hat{C}_{i,j} \leq \Gamma_{1-p} \left( \frac{1}{2}(|\mathcal{X}|-1)(|\mathcal{Y}| - 1),  \frac{1}{N \log(2)} \right)
    \label{eq:signi_condition}
\end{equation}
Where $\Gamma_{1-p}(\alpha, \beta)$ is the $(1-p)$ quantile of a Gamma distribution $\mathcal{G}(\alpha, \beta)$. 
If (\ref{eq:signi_condition}) is verified, the test accepts the null hypothesis, i.e. $A_m$ and $B_n$ are independent, and rejects it otherwise. Since we are performing multiple statistical hypothesis testings, the likelihood of incorrectly rejecting the null hypothesis increases. As in \cite{curme2015emergence} and \cite{fiedor2014information}, we compensate for that increase by applying the Bonferroni correction : we test each individual hypothesis at a significance level of $p / m$ where $m=n^2$ is the total number of tests. \\

We denote by $\mathbf{\Tilde{C}} \in (\mathbb{R}+)^{n \times n}$ the matrix for which entry $(i, j)$ is equal to $\hat{C}_{i,j}$ if it is significant, otherwise 0. The matrix $\mathbf{\Tilde{C}}$ can be seen as a weighted adjacency matrix of a directed graph. We decide to symmetrize the matrix $\mathbf{\Tilde{C}}$ so that we can use most of existing node embedding algorithms: \\

\begin{equation}
    \label{eq:symm_}
    \mathbf{\Tilde{C}} \leftarrow \frac{\mathbf{\Tilde{C}} + \mathbf{\Tilde{C}}^t}{2}
\end{equation}

%% file: sections/deepNF.tex
In this section, we introduce the framework of \textit{deepNF} introduced by \citeauthor{gligorijevic2018deepnf} \cite{gligorijevic2018deepnf}. \textit{DeepNF} learns low-dimensional embedding of nodes, shared across several graphs. We consider a set of $N$ graphs which are represented by their binary adjacency matrices $ \{\mathbf{A^{(i)}}\}_{i=1}^{N}  \triangleq \{\mathbf{A}^{(1)}, ..., \mathbf{A}^{(N)}\}$, i.e. $A_{i,j}^{(l)} = 1$ if nodes $i$ and $j$ are connected in graph $l$ and 0 otherwise. Their approach is composed of three steps: first converting each graph into a high-quality vector representation with the \textit{Random Walk with Restarts} (RWR) method, then constructing from each RWR matrix a \textit{Positive Pointwise Mutual Information} (PPMI) matrix capturing structural information, and finally fusing PPMI matrices by using an \textit{AutoEncoder} like model. The entire methodology is schematized in figure 1. \\

\begin{figure}
    \centering
    \includegraphics[width=0.9\linewidth]{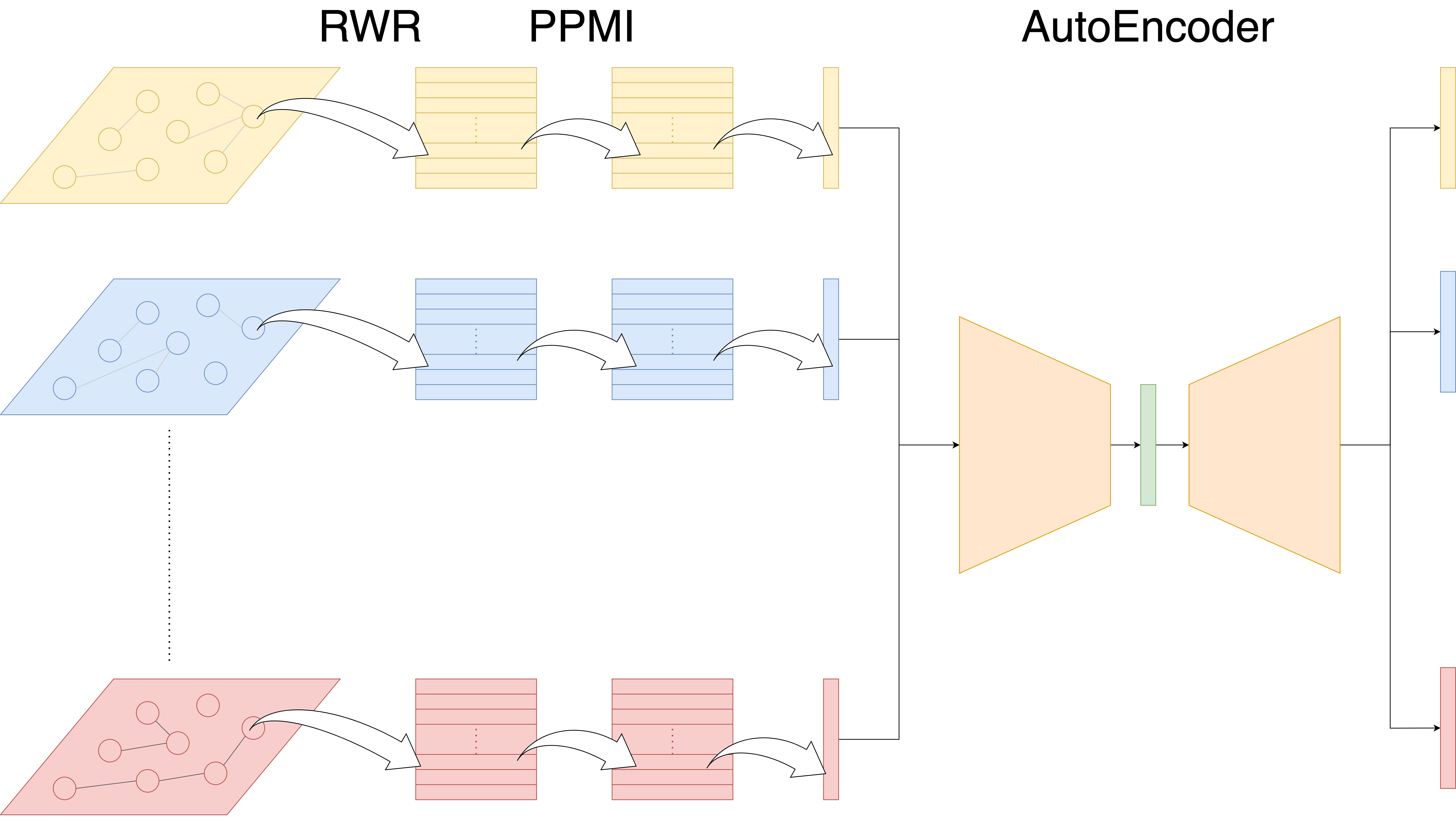}
    \label{fig:deepNF}
    \caption{Deep Network Fusion}
\end{figure}

(1) The \textit{Random Walk} (RW) method is an effective way to transform an unweighted graph into node representations that capture the topological structure of each node. Let $\mathbf{A} \in [0, 1]^{n \times n}$ be a binary adjacency matrix on a set of $n$ nodes $\mathcal{N}$. We make the following assumption: the node $i$ is self connected, i.e. $A_{i, i} = 1$, if and only if it is isolated, i.e. $\forall j \neq i, \ A_{i, j} =0$. Given a starting node $i_0 \in \mathcal{N}$ and a length $L \in \mathbb{N}^{*}$, a random walk is a path $i_0 \rightarrow i_1 \rightarrow ... \rightarrow i_{L-1}$, where for all $j \in \{1, ..., L-1\}$, $i_j \in \mathcal{N}$ and $i_j$ is uniformly sampled among the neighbours of $i_{j-1}$. A \textit{Random Walk with Restart} (RWR) allows at each time step to return to $i_0$ with a probability $1-\alpha$, where $\alpha \in [0, 1)$. RWR can be formulated by the recurrence relation of equation \ref{eq:random_walk}. 

\begin{equation}
    \mathbf{p}_i^{(t)} = \alpha \times  \mathbf{p}_i^{(t-1)} \hat{\mathbf{A}} + (1-\alpha) \times \mathbf{p}_i^{(0)},
    \label{eq:random_walk}
\end{equation}

where for all $t$, $\mathbf{p}_i^{(t)}$ is a row vector of size $n$, whose $k$-th entry indicates the probability of reaching the node $k$ starting from node $i$ after $t$ steps and $\hat{\mathbf{A}}$ is the one-step probability transition matrix obtained from $\mathbf{A}$ by applying the standard row-wise normalization. In order to give a representation $\mathbf{v}_i \in \mathbb{R}^n$ for the node $i$, \citeauthor{gligorijevic2018deepnf} \cite{gligorijevic2018deepnf} adopt the method proposed by \citeauthor{cao2016deep} \cite{cao2016deep} given by the equation \ref{eq:repr_rwr}.
\begin{equation}
    \mathbf{v}_i = \sum_{t=1}^K \mathbf{p}_i^{(t)},
    \label{eq:repr_rwr}
\end{equation}
where $K$ is the total number of RWR steps. As a consequence, the representation $v_i$ of node $i$ accounts for the neighbourhood of node $i$ up to nodes that are distant of $K$ from $i$. Since $p^{(t)}_i$ is a probability vector, it is worth noticing that $\sum_{j=1}^n v_{i, j} = K$. Applying the RWR step for each graph results in $N$ representation matrices $\{ \mathbf{V}^{(1)}, ..., \mathbf{V}^{(N)} \}$ of size $\mathbb{R}^{n \times n}$ : 

\begin{equation}
    \forall l \in \{1, ..., N\}, \ \mathbf{V}^{(l)} = 
    \left(
    \begin{array}{c}
         \mathbf{v}_1^{(l)} \\
        \mathbf{v}_2^{(l)} \\
         \vdots \\
         \mathbf{v}_i^{(l)} \\
         \vdots \\
         \mathbf{v}_n^{(l)}
    \end{array}
    \right)
\end{equation}

As a consequence:

\begin{equation}
    \forall l \in \{1, ..., N\}, \ \forall i, j \in \{0, ..., n\}, \ V^{(l)}_{i,j} = \sum_{t=1}^K \mathbf{p}_{i, j}^{(t)}
\end{equation}

(2) Given a RWR matrix $\mathbf{V}^{(l)}$, we denote by $\Tilde{\mathbf{V}}^{(l)}$ the matrix $\mathbf{V}^{(l)}$ normalized:
\begin{equation}
    \Tilde{\mathbf{V}}^{(l)} = \frac{\mathbf{V}^{(l)} }{\sum_{q=1}^n \sum_{r=1}^n V^{(l)}_{qr}} = \frac{\mathbf{V}^{(l)} }{n \times K},
\end{equation}
$\Tilde{\mathbf{V}}^{(l)}$ can be seen as the matrix of probabilities of two variables $(X, Y)$ taking their values in $\{1, ..., n \}$, i.e. $p(X=i, \ Y=j) = \Tilde{\mathbf{V}}^{(l)}_{ij}$. It is straightforward that $p(X=i) = \sum_{j=1}^n \Tilde{\mathbf{V}}^{(l)}_{ij} = n^{-1}$ and $p(Y=j) = \sum_{i=1}^n \Tilde{\mathbf{V}}^{(l)}_{ij}$. We construct the corresponding PPMI matrix $\mathbf{P}^{(l)}$ as follows:

$$\forall i \in \{1, ..., n\}, \ \forall j \in \{1, ..., n\}$$

\begin{equation}
\begin{split}
    \mathbf{P}^{(l)}_{i, j}  & =   \max \left( 0, \log\left( \frac{p(X=i, \ Y=j)}{p(X=i) p(Y=j)} \right) \right) \\
& =   \max \left( 0, \log\left( \frac{\Tilde{\mathbf{V}}^{(l)}_{ij}}{\left( \sum_{j=1}^n \Tilde{\mathbf{V}}^{(l)}_{ij} \right) \left( \sum_{i=1}^n \Tilde{\mathbf{V}}^{(l)}_{ij} \right)} \right) \right) \\
& =   \max \left( 0, \log\left( \frac{ n \times \mathbf{V}^{(l)}_{ij}}{\sum_{q=1}^n \mathbf{V}^{(l)}_{qj}  } \right) \right).
\end{split}
\label{eq:PPMI}
\end{equation}
At the end of this step, each node $i$ is represented by $N$ row vectors $  \{\mathbf{P}^{(j)}_i\}_{j=1}^{N} \triangleq  \{\mathbf{P}^{(1)}_i, ..., \mathbf{P}^{(N)}_i \}$, where for all $l \in \{1, ..., N\}$, $\mathbf{P}_i^{(l)} \in \mathbb{R}^n$. \\

 (3). Let $\mathcal{X}$ and $\mathcal{Z}$ be two mathematical spaces. An \textit{AutoEncoder} is the combination of two parametrized functions $\Phi : \mathcal{X} \rightarrow \mathcal{Z}$ and $\Psi: \mathcal{Z} \rightarrow \mathcal{X}$, respectively the \textit{encoder} and the \textit{decoder}. The associated parameters are denoted by $\theta_{\Phi}$ and $\theta_{\Psi}$. Let $x \in \mathcal{X}$ be a sample, we define the embedding $z \in \mathcal{Z}$ and the reconstruction $\hat{x} \in \mathcal{X}$ of $x$ respectively by $z := \phi(x, \theta_{\Phi})$ and $\hat{x} := \Psi(z, \theta_{\Psi})$. If the dimension of $\mathcal{Z}$ is smaller than the dimension of $\mathcal{X}$, $z$ is a compressed representation of $x$.  In order to find a meaningful representation $z$, the objective is to minimize a reconstruction error between $x$ and $\hat{x}$. In the following, we build our encoder and decoder with neural networks called \textit{multilayer perceptron} \cite{ramchoun2016multilayer} (MLP).

\textit{Multilayer perceptron} refers to a feedforward neural network composed of multiple linear layers with non linear activation. A \textit{multilayer perceptron} can thus be described by its activation function and the ordered output dimensions of its linear layers, namely the latent dimensions.  An MLP \textit{encoder} (resp. \textit{decoder}) is a multilayer perceptron, whose latent dimensions are decreasing (resp. increasing) w.r.t. the depth. Next, we define the model used by \citeauthor{gligorijevic2018deepnf} \cite{gligorijevic2018deepnf} to fuse the different representations (found at the end of step 2) of each node. Given a  graph $l \in \{1, ..., N\}$, we compress the representation $\mathbf{P}_i^{(l)}$ of node $i$ into $\mathbf{z}^{(l)}_i$ using a MLP encoder $\mathrm{eMLP}^{(l)}$ (equation \ref{eq:encoding}).
    
\begin{equation}
    \forall j \in \{1, ..., N\}, \ \forall i \in \{1, ..., n\}, \ \mathbf{z}^{(l)}_i = \mathrm{eMLP}^{(l)}\left(\mathbf{P}^{(l)}_i \right)
    \label{eq:encoding}
\end{equation}

Next, we concatenate for each node $i$ its representations $(\mathbf{z}^{(l)}_i)_{1 \leq i \leq N}$ obtained previously (equation \ref{eq:concatenation}) and feed it to a MLP encoder eMLP in order to get the fused representation $\mathbf{z}_i$ of node $i$ (equation \ref{eq:common_encoding}). 

\begin{equation}
    \forall i \in \{1, ..., n \}, \ \mathbf{x}_i = \mathrm{Concat}\left(\mathbf{z}^{(1)}_i, ..., \mathbf{z}^{(N)}_i \right)
    \label{eq:concatenation}
\end{equation}
\begin{equation}
    \forall i \in \{1, ..., n \}, \ \mathbf{z}_i = \mathrm{eMLP}(\mathbf{x}_i)
    \label{eq:common_encoding}
\end{equation}

Then, $\mathbf{z}_i$ is fed to a MLP decoder dMLP and give the decompressed representation $\mathbf{\hat{x}}_i$ of $\mathbf{z}_i$. $\mathbf{\hat{x}}_i$ is then split into $N$ non overlapping vectors $\{ \mathbf{\hat{z}}^{(1)_i}, ..., \mathbf{\hat{z}}^{(N)}_i \}$. Finally, each vector $\mathbf{\hat{z}}^{(l)}_i$ is fed to a MLP decoder $\mathrm{dMLP}^{(l)}$ whose last latent dimension is equal to the dimension of $\mathbf{P}_i^{(l)}$. As a consequence, if we denote by $\mathbf{\hat{P}}_i^{(l)}$ the output of $\mathrm{dMLP}^{(l)}$, $\mathbf{\hat{P}}_i^{(l)}$ has the same size than $\mathbf{P}_i^{(l)}$. This network aims at finding a meaningful representation $\mathbf{z}_i$ of node $i$, and this representation is validated and refined by attempting to regenerate the inputs $\{ \mathbf{P}^{(1)}_i\}_{i=1}^{N} \triangleq \{\mathbf{P}^{(1)}_i, ..., \mathbf{P}^{(N)}_i \}$  from $\mathbf{z}_i$. For this reason, we try to find the optimal parameters that minimize the reconstruction loss $\mathcal{L}$ between each original and reconstructed PPMI matrix (equation \ref{eq:reconstruction_loss}).

\begin{equation}
    \mathcal{L} \triangleq \frac{1}{N} \sum_{l=1}^N L(\mathbf{\hat{P}}^{(l)}, \mathbf{P}^{(l)})
    \label{eq:reconstruction_loss}
\end{equation}

where $L$ is the mean squared error. This loss function can be optimized by a standard back-propagation algorithm. After the training of the model is done, we extract for each node $i$ its low-dimensional feature vector $\mathbf{z}_i$, in the following called embedding of  asset $i$.

%% file: sections/deep_fusion_lead_lag_networks.tex
Let $ \{ (d_i, T_i) \}_{i=1}^{N} \triangleq \{(d_1, T_1), (d_2, T_2), ...., (d_N, T_N)\}$ be a set of tuples of $\mathbb{R} \times \mathbb{N}$ composed of a sampling period $d$ and a lag $T$. For all $l \in \{1, ..., N\}$, we denote by $\mathbf{\Tilde{C}}^{(l)}$ the $T^{(l)}$ lead-lag network obtained from the series of prices $ \{ \mathbf{P}^{i} \}_{i=1}^{N} \triangleq \{\mathbf{P}^{(1)}, ..., \mathbf{P}^{(N)}\}$ sampled with the frequency $1 / d^{(l)}$. It is worth noticing that two distinct value of $d$ lead to two different matrices of returns (equation \ref{eq:return_matrix}) because the series of prices are not sampled with the same frequency. In the following, we will denote by $\mathbf{R}^{(d)}$ the matrix of returns derived from the sampling period $d$ when the confusion is possible. The framework of deepNF enables to find a representation of each asset from $ \{\mathbf{\Tilde{C}}^{(i)}\}_{i=1}^{N} \triangleq \{\mathbf{\Tilde{C}}^{(1)}, ...,  \mathbf{\Tilde{C}}^{(N)} \}$. These representations have the particularity to contain information from multiple multi-scale non-linear lagged relationships between assets. It is worth noticing that this approach is not only limited to lead-lag graphs, any kind of graph whose nodes are assets can be incorporated into the representation learning process, for example the  network of cryptocurrency sectors or the network of cryptocurrency investment firms. 

\subsection{Dynamic Lead-lag Graphs}

Since we want to detect new relations between assets, we have to reevaluate the relations between assets, and then the lead-lag graphs, and to update the representation of each asset on a regular basis. Concretely we choose $m$ distinct dates among the $p+1$ possible dates, it is equivalent to choose a  non decreasing injective function $\chi : \{0, ..., m-1\} \rightarrow \{0, ..., p\}$. Given an index $k \in \{0, ..., m-1\}$ and window size $w \in \mathbb{N}^*$, the date of the sample end is $t_{\chi(k)}$, in addition, we keep only data in a lookback window $[t_{\chi(k)-w+1}, t_{\chi(k)}]$ in order to catch recent dynamics. It concretely means to keep only the rows in $\mathbf{R}$ corresponding to the dates $t_{\chi(k) - w + 1}, t_{\chi(k) - w + 2}, ..., t_{\chi(k)}$, i.e. $\mathbf{R}_{[\chi(k) - w + 1, \chi(k)]}$. From $\mathbf{R}_{[\chi(k) - w + 1, \chi(k)]}$, we compute the $N$ lead-lag graphs $\{\mathbf{\tilde{C}^{(i)}} (t_{\chi(k)})\}_{i=1}^{N}$ and their binary adjacency matrices $\{\mathbf{A^{(i)}}(t_{\chi(k)})\}_{i=1}^{N}$. \textit{DeepNF} is then trained to fuse the set of adjacency matrices obtained over all dates $\{ \{\mathbf{A^{(i)}}(t_{\chi(k)})\}_{i=1}^N \}_{k=0}^{m-1}$. Finally, embeddings $\{ \{ z_j(t_{\chi(k)}) \}_{j=1}^{n} \}_{k=0}^{m-1}$ are extracted from the encoder. Given an asset $j \in \{1, ..., n\}$, $m$ embeddings are available, one for each date $\{t_{\chi(k)}\}_{k=0}^{m-1}$. As a consequence, our framework allows the considered assets to have a time-varying representation.

\begin{algorithm}[]
 \KwData{$\{ P_j \}_{j=1}^n$, $\{ (d_i, T_i) \}_{i=1}^{N}$, $w$, $\chi$}
 \KwResult{Dynamic embeddings $\{\{ z_j(t_{\chi(k)}) \}_{j=1}^n \}_{k=0}^{m-1}$}
 \textit{Initialization}\;
 \For{$l=1,..., N$}{
 Compute the matrix of returns $\mathbf{R}^{(d_l)}$ with the frequency $d_l$
 }
 \textit{Training}\;
 \For{$k=1, ..., m$}{
 $t \leftarrow \chi(k)$\;
 \For{$l=1,..., N$}{
 $(d, T) \leftarrow (d_l, T_l)$\;
 $R \leftarrow \mathbf{R}^{d}_{t-w+1:t, :}$\;
 Construct the $T$ lead-lag graph $\tilde{C}^{(l)}(t)$ from $R$\;
 Derive the binary adjacency matrix $A^{(l)}(t)$ from $\tilde{C}^{(l)}(t)$
 }
 }
 Train DeepNF to fuse $\{ \{ A^{(l)}(\chi(k)) \}_{l=1}^N \}_{k=0}^{m-1}$\;
 \textit{Ending}\;
 Extract $\{\{ z_j(t_{\chi(k)}) \}_{j=1}^n \}_{k=0}^{m-1}$ from the encoder of DeepNF
 \vspace*{0.3cm}
 \caption{Dynamic Deep Fusion}
\end{algorithm}

%% file: sections/material.tex
\subsection{Lead-lag Graphs}
\begin{figure}
\centering
\begin{minipage}{.40\textwidth}
  \centering
  \includegraphics[width=.8\linewidth]{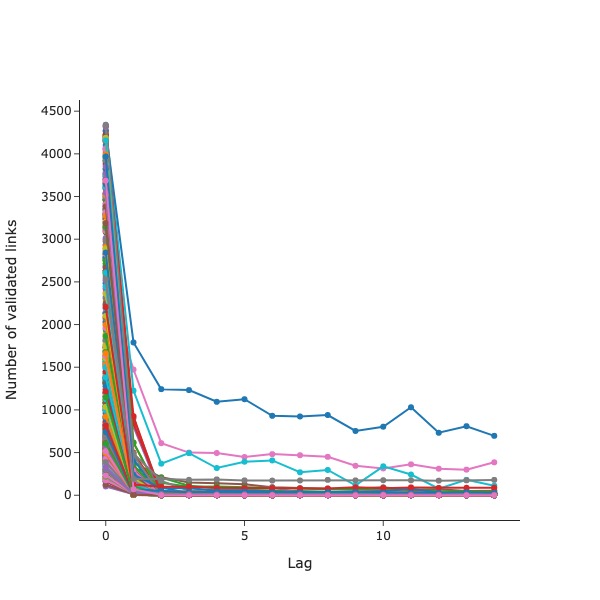}
  \caption{Sampling period : $d=1$ minute. Number of validated links for different lags $T$. One curve per date.}
  \label{fig:number_val_link_1minute}
\end{minipage}%
\hspace{0.2cm}
\begin{minipage}{.40\textwidth}
  \centering
  \includegraphics[width=.8\linewidth]{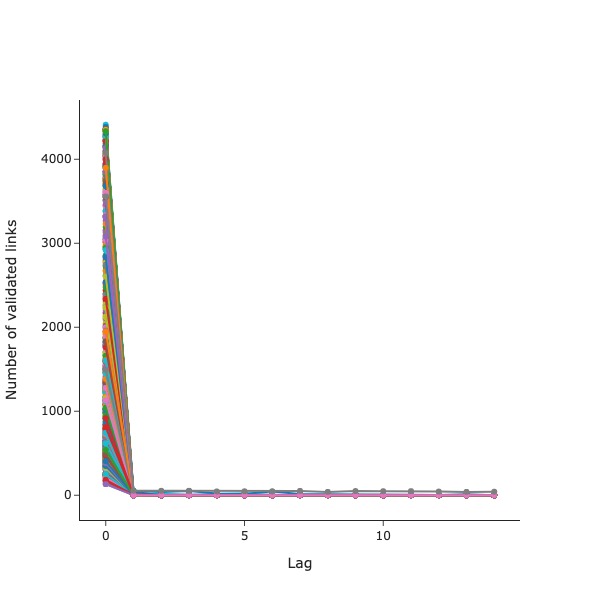}
  \caption{Sampling period : $d=5$ minutes. Number of validated links for different lags $T$. One curve per date.}
  \label{fig:number_val_link_5minutes}
\end{minipage}
\end{figure}

{
In order to find mutual information-based lagged relationships between cryptoassets, we have downloaded the quotes on a minute-by-minute basis of all assets present in the weekly ranking on market capitalization established by CoinMarketCap (\url{https://coinmarketcap.com/fr/historical/}) on the 13th October 2019. We have filtered out assets that were not listed on Binance (\url{www.binance.com}) at this date and quotes\footnote{The quote considered is \textit{token}/USDT.} of the resulting $n=69$ assets\footnote{The remaining tokens are: ADA, ALGO, ANKR, ATOM, BAND, BAT, BEAM, BNB, BTC, BTT, BUSD, CELR, CHZ, COCOS, COS, CVC, DASH, DENT, DOGE, ENJ, EOS, ETC, ETH, FET, FTM, FUN, GTO, HBAR, HOT, ICX, IOST, KAVA, KEY, LINK, LTC, MATIC, MFT, MITH, MTL, NANO, NEO, NKN, NULS, OMG, ONE, ONG, ONT, PERL, QTUM, REN, RVN, STX, TFUEL, THETA, TOMO, TRX, TUSD, USDC, VET, WAN, WAVES, WIN, XLM, XMR, XRP, XTZ, ZEC, ZIL, ZRX} have been downloaded from the Binance API (\url{https://binance-docs.github.io}). The data cover 705 days between  12th September 2019 and  7th  November  2021. Each sample corresponds to $n$ series of quotes from a lookback window of size $w=1440$ minutes, i.e. 24 hours of data, ending at the end of a day.  As a result there is no overlap between two consecutive samples. We compute from the series of quotes the series of log-returns using two sampling periods : 1 minute and 5 minutes. For the purpose of estimating mutual information, these log-returns are sample- and asset-wise discretized into 4 distinct states. As in \cite{fiedor2014information}, the states represent equal parts, therefore each state is assigned the same number of data points. We have set the uncorrected p-value $p$ (equation \ref{eq:signi_condition}) equal to 0.01, using the Bonferroni correction this p-value is set to $0.01 / 69^2$. On figures \ref{fig:number_val_link_1minute} and \ref{fig:number_val_link_5minutes} we plot for each sample, i.e. for each date,  the number of validated mutual information-based links for the first 15 lags for log-returns sampled with periods 1 minute and 5 minutes respectively. In addition, we report on tables \ref{tab:statistics_1m} and \ref{tab:statistics_5min} some lag-wise statistics for the sample periods 1 minutes and 5 minutes respectively. The lead-lag graphs themselves for different values of $T$, three distinct dates and both sampling periods are shown on figures \ref{fig:lag_net_1_0_08-12-2019}::\ref{fig:lag_net_5_2_2021-09-08}. We recall that nodes are assets, and that an edge from an asset to another is drawn if a significant lagged relationship has been found between those assets.
}

\begin{table*}[t]
    \centering
    \begin{tabular}{c|ccccccccccccccc}
        Lag & 0 & 1 & 2 & 3 & 4 & 5 & 6 & 7 & 8 & 9 & 10 & 11 & 12 & 13 & 14 \\
        \hline
        Min & 103 & 3 & 0 & 0 & 0 & 0 & 0 & 0 & 0 & 0 & 0 & 0 & 0 & 0 & 0 \\
        Quantile 25 \% & 601 & 15 & 1 & 0 & 0 & 0 & 0 & 0 & 0 & 0 & 0 & 0 & 0 & 0 & 0 \\
        Median & 1332 & 20 & 2 & 1 & 1 & 1 & 1 & 1 & 1 & 1 & 0 & 0 & 0 & 0 & 0 \\
        Quantile 75 \% & 2881 & 35 & 4 & 3 & 2 & 2 & 2 & 2 & 2 & 1 & 2 & 1 & 1 & 1 & 1 \\
        Max & 4341 & 1790 & 1243 & 1232 & 1095 & 1125 & 931 & 923 & 940 &  753 & 803 & 1031 & 732 & 809 & 696 \\
    \hline
    \end{tabular}
    \caption{Sampling period : $d=1$ minute. Statistics on the number of validated links for different lags $T$. }
    \label{tab:statistics_1m}
\end{table*}

\begin{table*}[t]
    \centering
    \begin{tabular}{c|ccccccccccccccc}
        Lag & 0 & 1 & 2 & 3 & 4 & 5 & 6 & 7 & 8 & 9 & 10 & 11 & 12 & 13 & 14 \\
        \hline
        Min & 133 & 0 & 0 & 0 & 0 & 0 & 0 & 0 & 0 & 0 & 0 & 0 & 0 & 0 & 0 \\
        Quantile 25 \% & 649 & 1 & 0 & 0 & 0 & 0 & 0 & 0 & 0 & 0 & 0 & 0 & 0 & 0 & 0 \\
        Median & 1339 & 2 & 0 & 0 & 0 & 0 & 0 & 0 & 0 & 0 & 0 & 0 & 0 & 0 & 0 \\
        Quantile 75 \% & 3037 & 3 & 1 & 0 & 0 & 0 & 0 & 0 & 0 & 0 & 0 & 0 & 0 & 0 & 0 \\
        Max & 4441 & 56 & 53 & 53 & 45 & 47 & 45 & 52 & 50 &  40 & 48 & 49 & 44 & 40 & 42 \\
    \hline
    \end{tabular}
    \caption{Sampling period : $d=5$ minute. Statistics on the number of validated links for different lags $T$. }
    \label{tab:statistics_5min}
\end{table*}

\subsection{Deep Fusion of Lead-lag Graphs}

\begin{figure}[h]
    \centering
    \includegraphics[width=0.7\columnwidth]{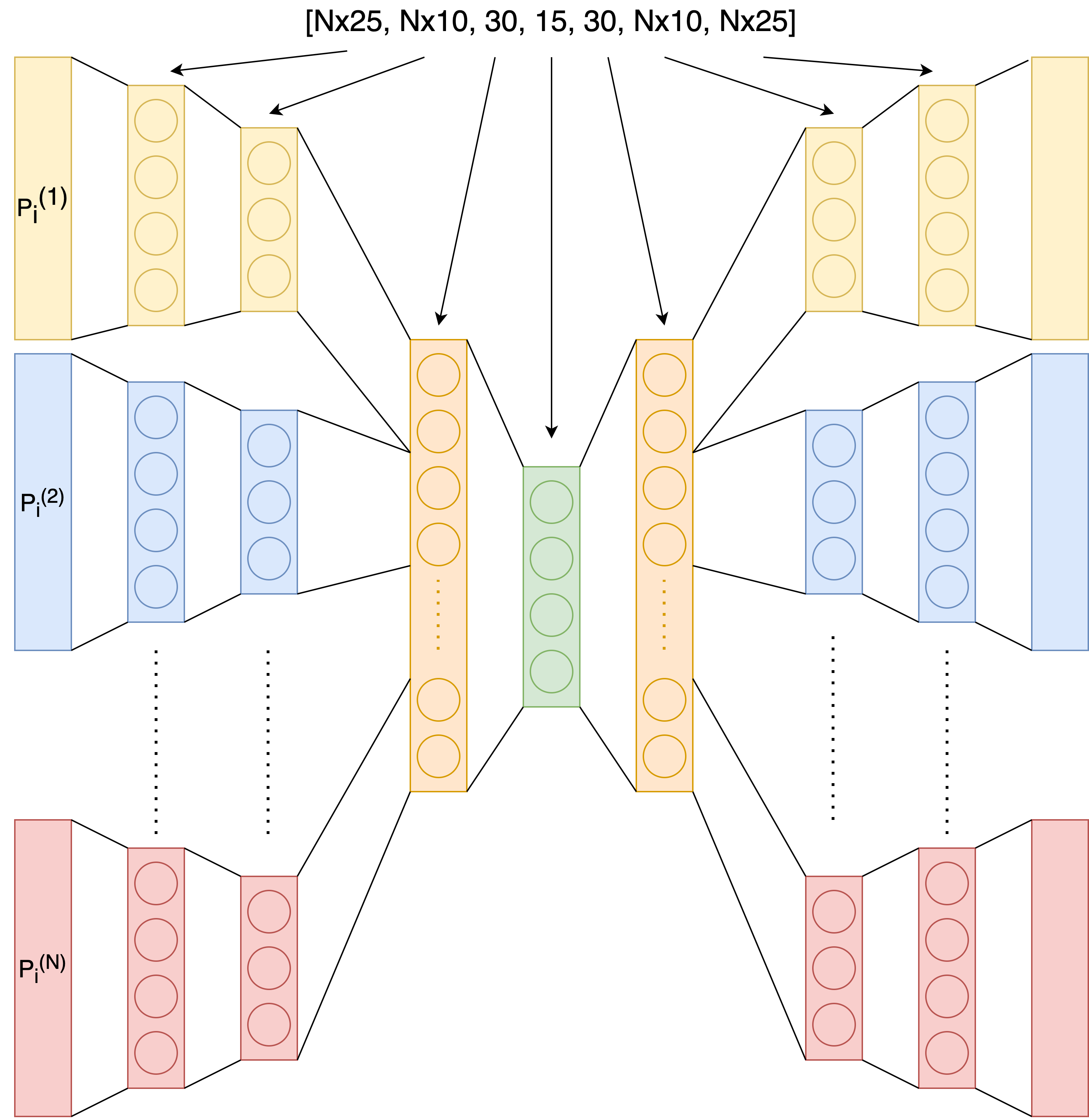}
    \caption{AutoEncoder}
    \label{fig:autoencoder}
\end{figure}

In the following, for both sampling periods, we will only consider the lead-lag networks obtained for the three first lags. We compute the first representation of each node by applying the first two steps of the deepNF approach. The model used in the third step is an \textit{AutoEncoder} in which we choose the ReLU function as activation function and the output dimensions are $[N \times 25, N \times 10, 30, 15, 30, N \times 10, N \times 25, N \times 100$, where $N=6$. The model is schematized on figure \ref{fig:autoencoder}. The data is split into a train (70\%) and a validation (30\%) set. We use the mean square error (MSE) as reconstruction error, and train the model by the algorithm Adam with a learning rate set equal to $0.001$. The training lasts at most 500 epochs, and is stopped if overfitting is detected in the evolution of validation loss. 

\subsection{Asset Representation}

\begin{figure}[h]
    \centering
    \includegraphics[width=0.8\columnwidth]{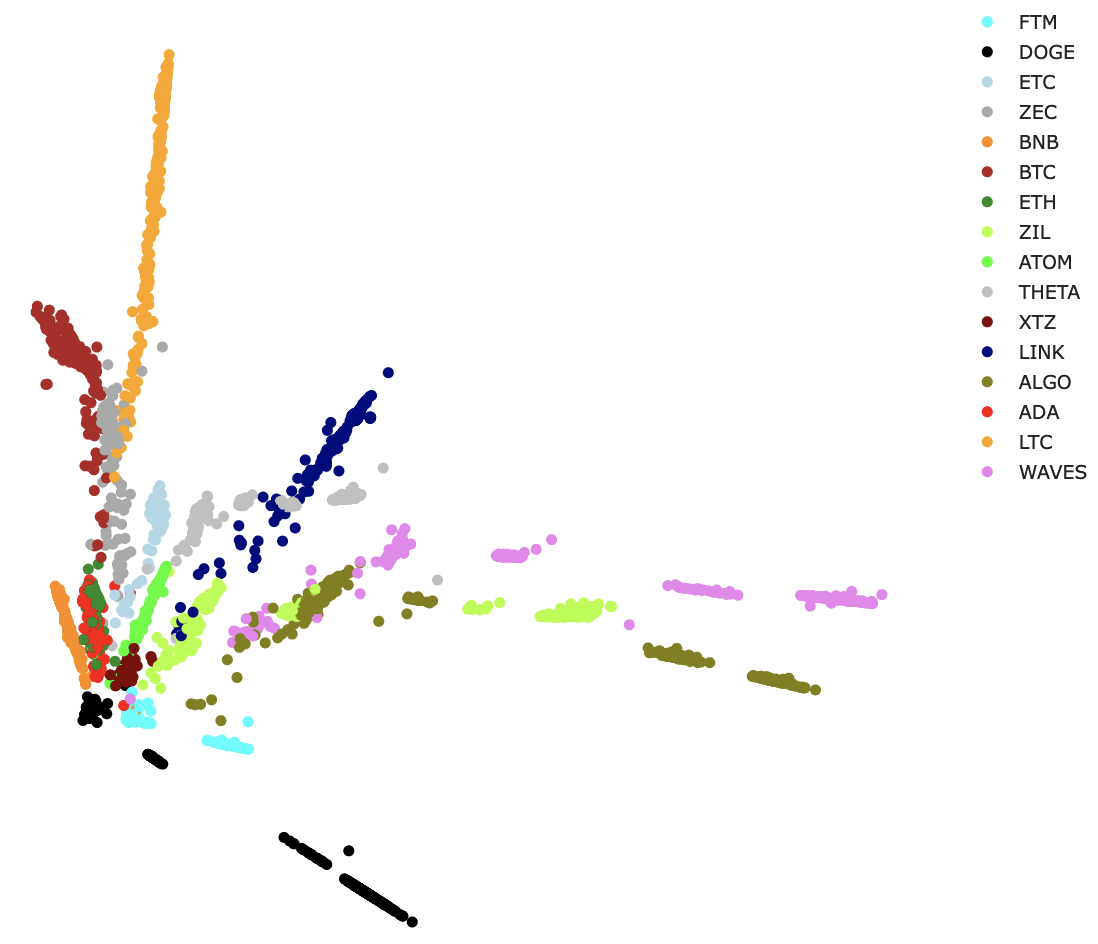}
    \caption{2D PCA projection of embeddings for the different dates of a subset of the studied assets.}
    \label{figure:PCA_top_coins}
\end{figure}

At the end of the training, the embeddings are extracted from the encoder. Given an asset, one embedding per sample / day is available. For a given asset $i$, we will denote by $z_i(t)$ the embedding of $i$ at date $t$. From all those embeddings, we train a model of \textit{Principal Component Analysis} (PCA) with two components. We plot the projected embeddings of a subset of 16 assets on figure \ref{figure:PCA_top_coins}. 

\begin{figure*}[]
\centering
\begin{minipage}{.60\columnwidth}
  \centering
  \includegraphics[width=.8\columnwidth]{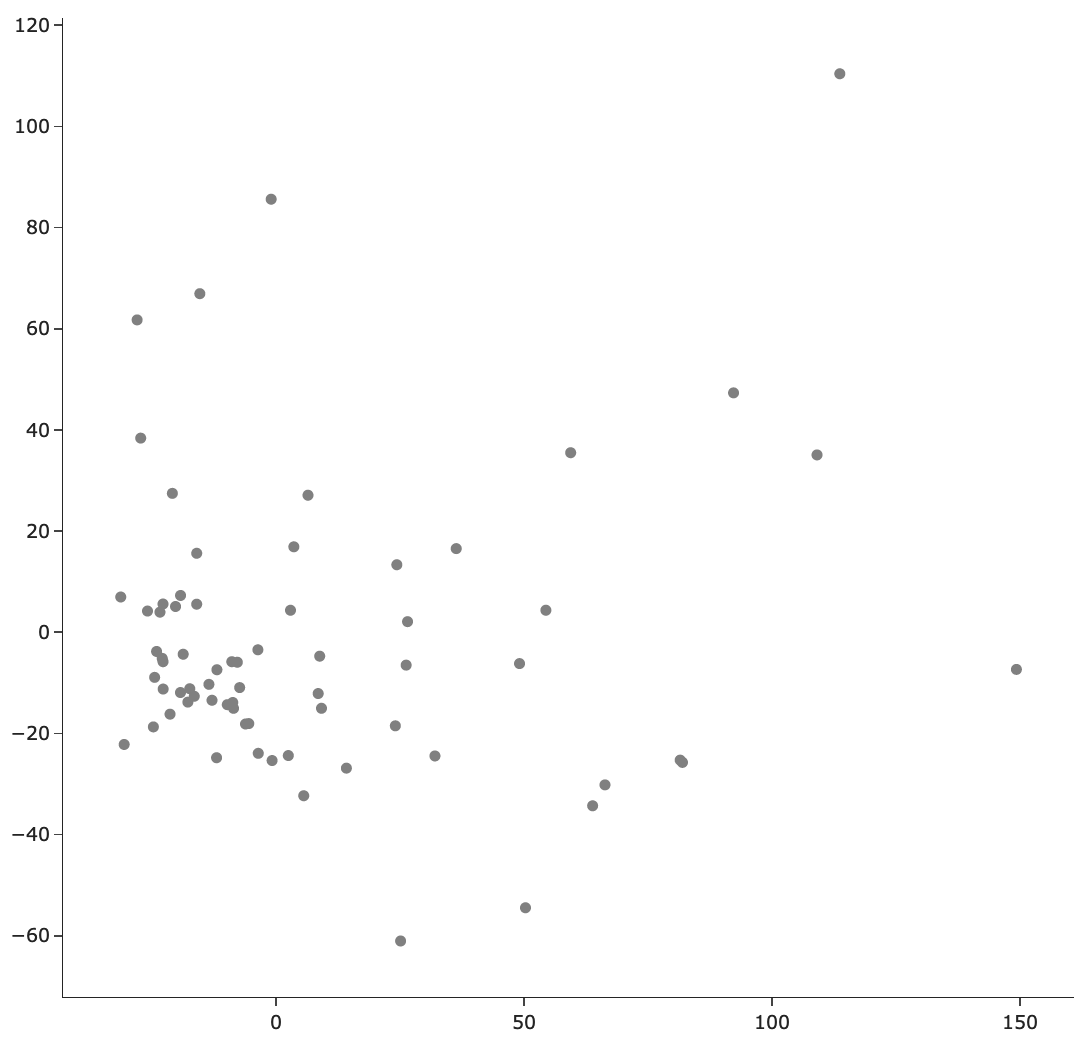}
  \caption{2D PCA projection of the embeddings on the 8th December of 2019.}
  \label{fig:pca_08_12_2019}
\end{minipage}%
\hspace{0.1cm}
\begin{minipage}{.60\columnwidth}
  \centering
  \includegraphics[width=.8\columnwidth]{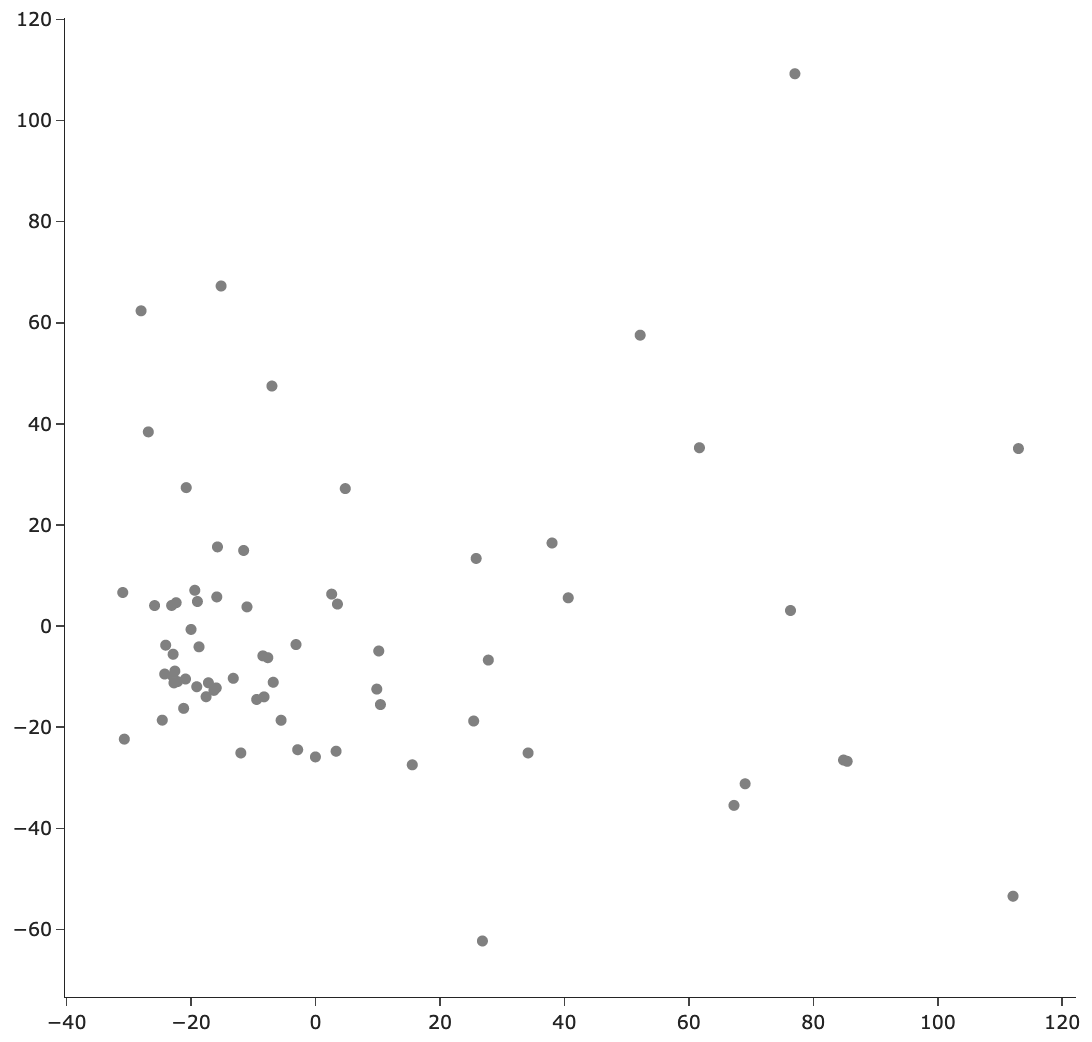}
  \caption{2D PCA projection of the embeddings on the 8th August of 2020.}
  \label{fig:pca_08_08_2020}
\end{minipage}%
\hspace{0.1cm}
\begin{minipage}{.60\columnwidth}
  \centering
  \includegraphics[width=.9\columnwidth]{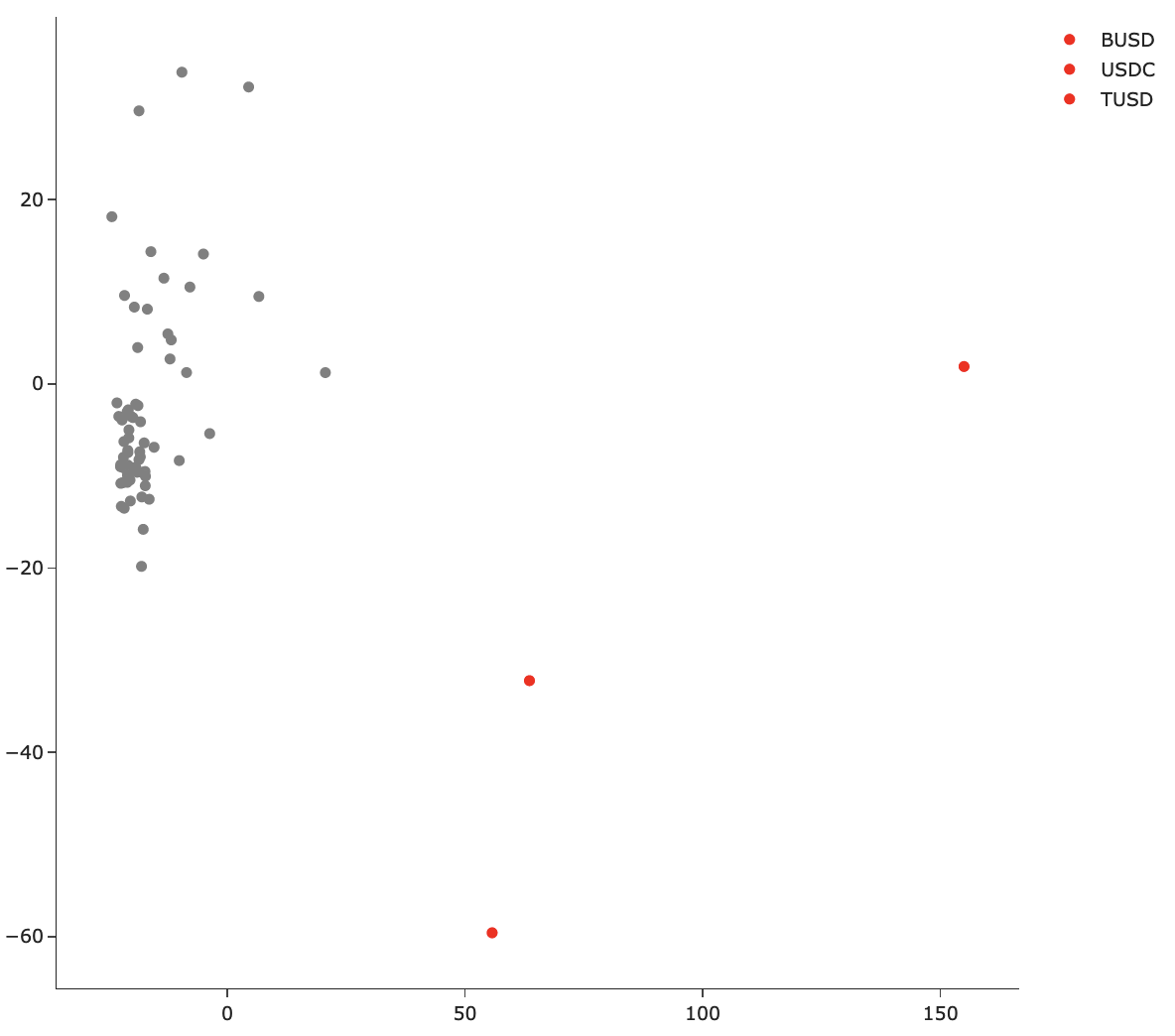}
  \caption{2D PCA projection of the embeddings on the 8th September of 2021.}
  \label{fig:pca_2021-09-08}
\end{minipage}
\end{figure*}

We also plot the projected embeddings of the the whole set of assets for the dates 8th December 2019, 8th August 2020 and 8th September 2021 on figures \ref{fig:pca_08_12_2019}, \ref{fig:pca_08_08_2020} and \ref{fig:pca_2021-09-08}, respectively. 

\begin{figure}
    \centering
    \includegraphics[width=0.9\columnwidth]{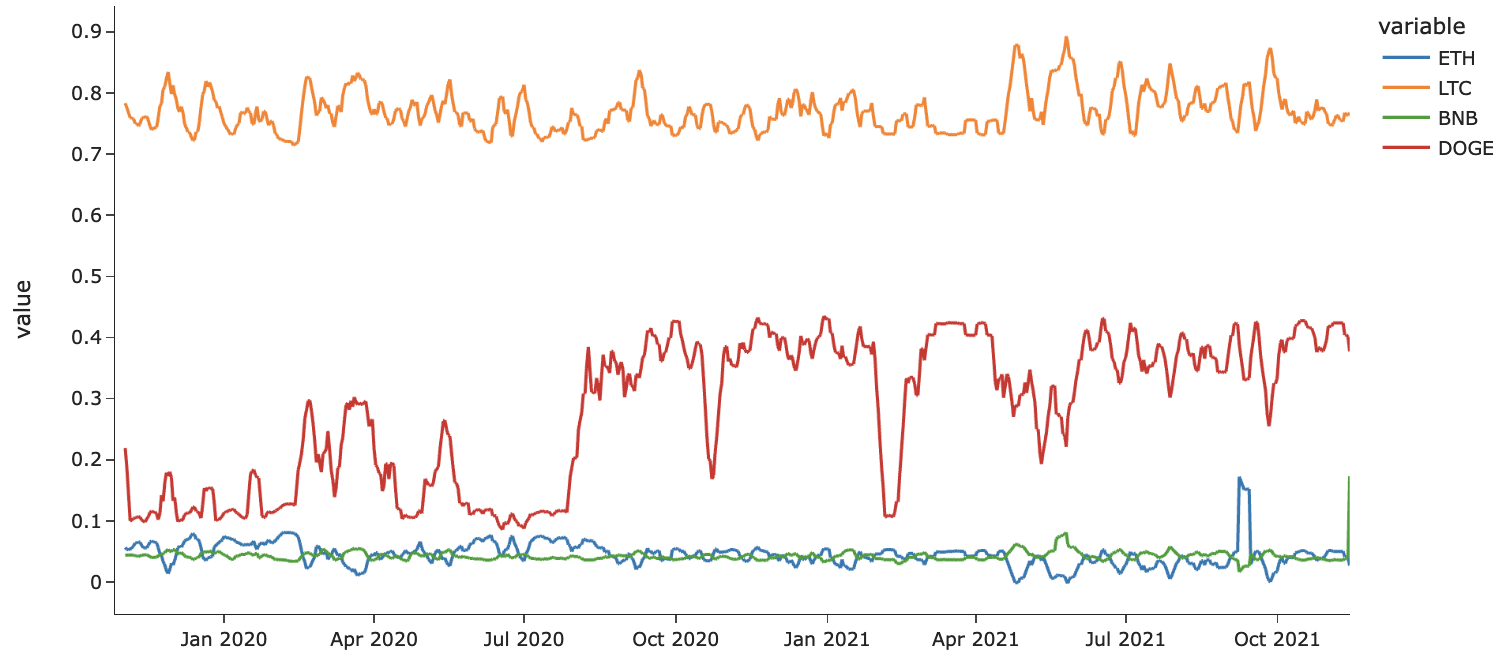}
    \caption{Evolution of the similarity between the token BTC and the tokens ETH, LTC, BNB et DOGE.}
    \label{fig:evolution_sim}
\end{figure}

Finally, a possible application of those representations is to monitor the temporal evolution of the similarity $S$ (equation \ref{eq:cossim}) between two assets. We plot on figure \ref{fig:evolution_sim} the evolution of the similarity between the asset BTC and assets ETH, LTC, BNB and DOGE.

\begin{equation}
    \forall i, j \in \{1, ..., n\}, \ S_{\cos}(\mathbf{z}_i(t), \mathbf{z}_j(t)) \triangleq \frac{ \bra{\mathbf{z}_i(t)} \ket{ \mathbf{z}_j(t)} }{||\mathbf{z}_i(t)||||\mathbf{z}_j(t)||}
    \label{eq:cossim}
\end{equation}

%% file: sections/discussion.tex
Figures \ref{fig:number_val_link_1minute}, \ref{fig:number_val_link_5minutes} and tables \ref{tab:statistics_1m} and \ref{tab:statistics_5min} demonstrate that a significant number of  synchronous and asynchronous relations exist between assets. The number of asynchronous relations rapidly decreases to 0 when the lag becomes large. Three dates are exception : the 19th May of 2021, the 12th March of 2020 and the 7th September of 2021, which all correspond to large bearish moves in the cryptocurrency markets. According to tables \ref{tab:statistics_1m}  and \ref{tab:statistics_5min}, there are a larger number of lagged relationships on a minute-to-minute basis than on a basis of 5 minutes. We deduce from the figures \ref{fig:lag_net_1_0_08-12-2019}::\ref{fig:lag_net_5_2_2021-09-08} that the graphs are significantly changing from one sample / day to another, and are unequally dense. In particular the graphs corresponding to the day of the large bearish movement of the 8th September of 2021 is extremely dense.

On figure \ref{figure:PCA_top_coins}, each asset seems to occupy it own area localized near a line in the plane. We notice that the central area $(0, 0)$ concentrates a large number of embeddings. Figures \ref{fig:pca_08_12_2019}, \ref{fig:pca_08_08_2020} and \ref{fig:pca_2021-09-08} seem to indicate that embeddings are concentrated during days of large moves (8th September 2021) and are more smeared during normal days (8th December 2019 and 8th August 2020). Stablecoins USDC, USDT and BUSD stand out from the others on figure \ref{fig:pca_2021-09-08}, it is however not surprising because they are not extensively impacted by large market moves. In a further study it would be interesting to study the relationship between the market return and the discrepancy of the embedding space for a given day. 

A similarity measure between two assets may be extremely useful, for example to minimize the risk of a portfolio. Similar assets can also be good candidates for pair trading strategies. On figure \ref{eq:cossim}, we can notice that the similarity between LTC and BTC is high and stable, it may be because LTC is a fork of BTC. Similarity between BTC and DOGE became only important since August 2020. In contrary, similarity between BTC and ETH or BNB remains globally weak.

%% file: conclusion.tex
In this paper, we have proposed a new approach to compute dynamic asset representations, which take into account the synchronous and asynchronous relations that may exist within a basket of assets. At the same time we have demonstrated the existence of those relations for sampling periods 1 minute and 5 minutes. In addition, it has been found that those relations were time-varying. At this point we consider two extensions of this work. First, it would be interesting that our approach supports a possibly time varying asset universe. In fact, cryptocurrencies is a relatively new sector, that is rapidly changing : new projects, and then assets, emerge every day. 
Secondly, from the computed dynamic embeddings, we can derive a dynamic similarity matrix by using the similarity defined by equation \ref{eq:cossim}. It would be interesting to determine if those matrices can be used to construct a risk-diversified portfolio.

%% file: sections/appendix.tex
\begin{figure*}[]
\centering
\begin{minipage}{.60\columnwidth}
  \centering
  \includegraphics[width=.8\columnwidth]{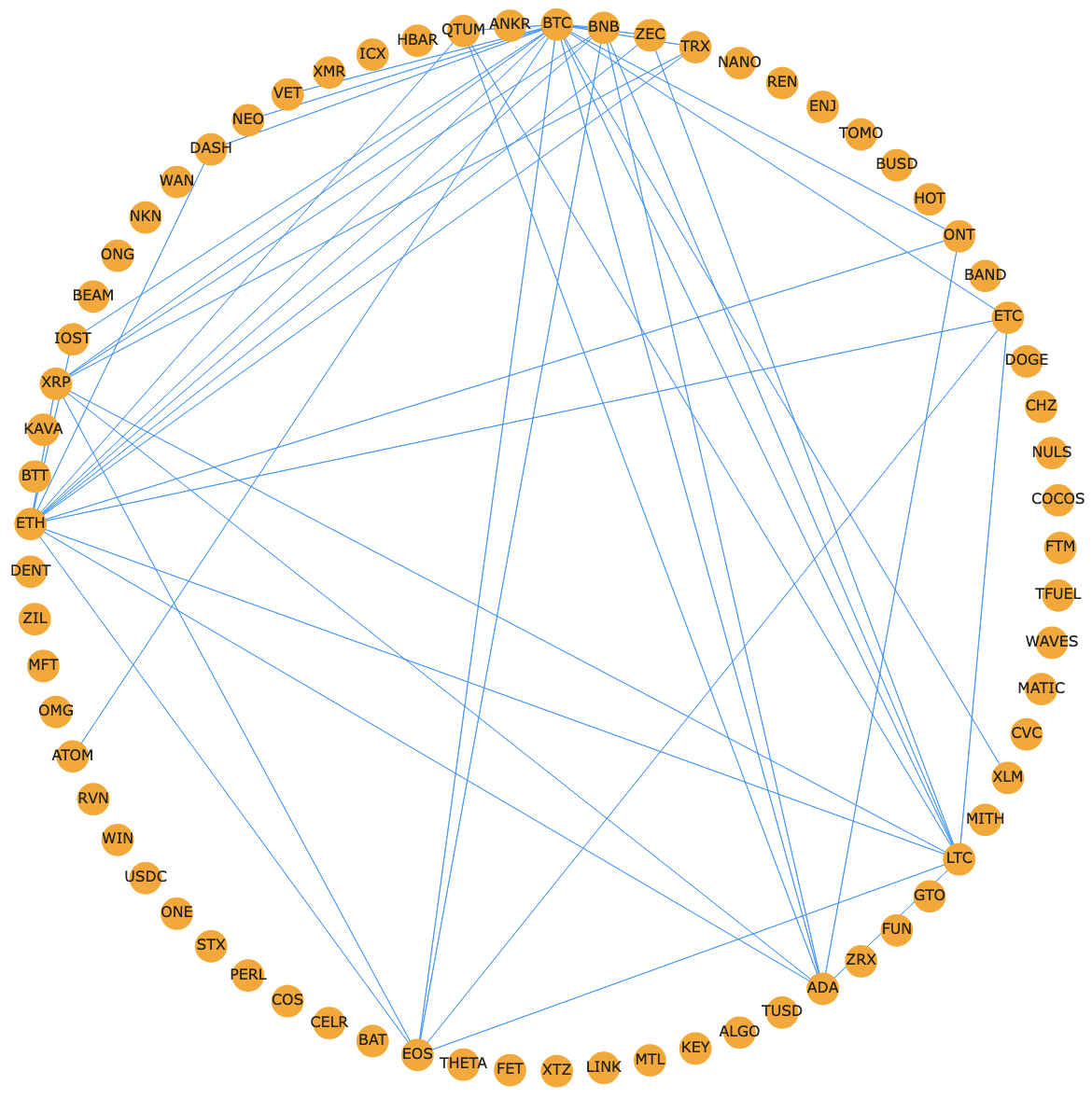}
  \caption{08-12-2019. Sampling period : $d=1$ minute. Lead-lag graph for $T = 0d$.}
  \label{fig:lag_net_1_0_08-12-2019}
\end{minipage}%
\hspace{0.1cm}
\begin{minipage}{.60\columnwidth}
  \centering
  \includegraphics[width=.8\columnwidth]{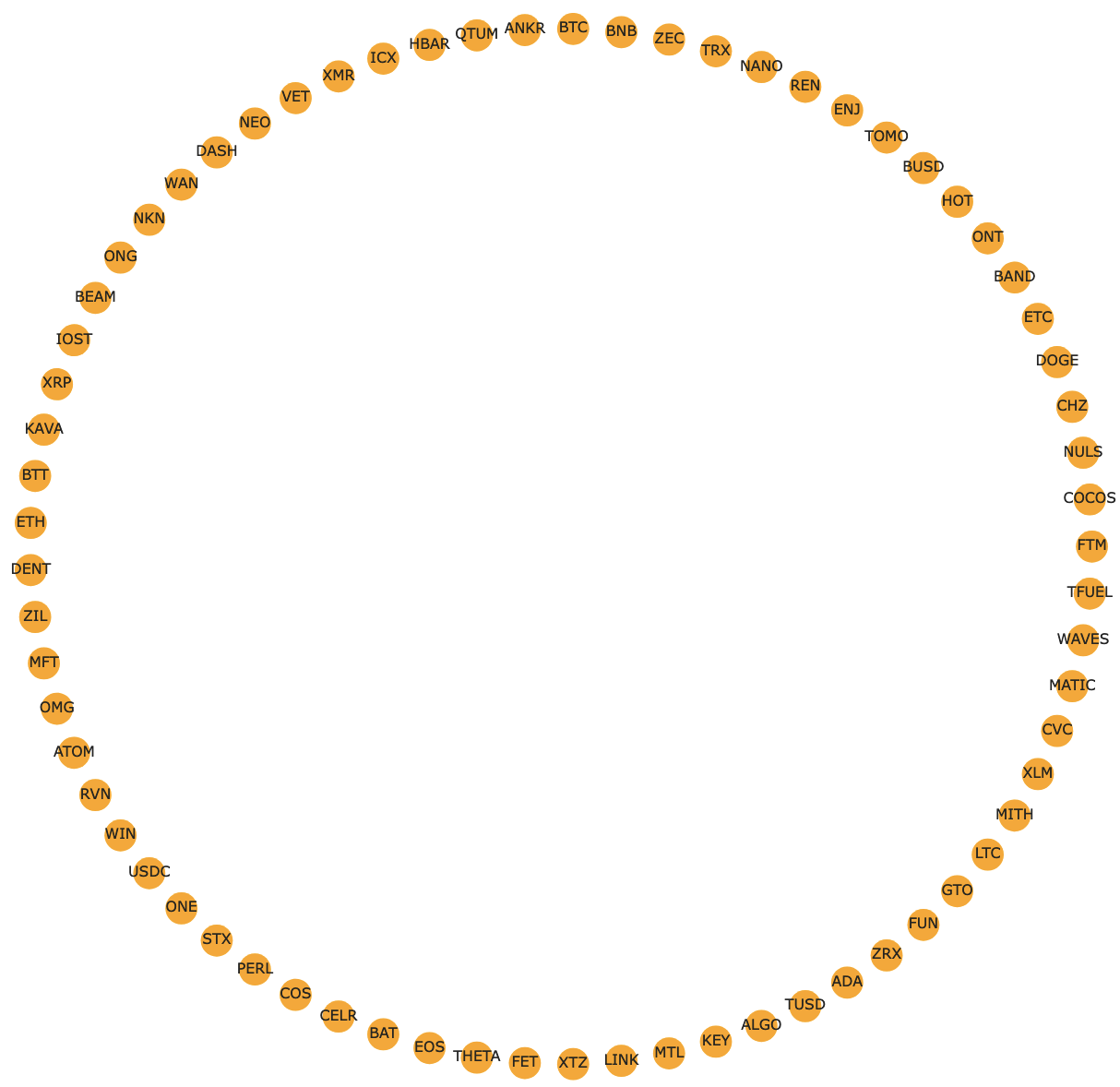}
  \caption{08-12-2019. Sampling period : $d=1$ minute. Lead-lag graph for $T = 1d$.}
  \label{fig:lag_net_1_1_08-12-2019}
\end{minipage}%
\hspace{0.1cm}
\begin{minipage}{.60\columnwidth}
  \centering
  \includegraphics[width=.8\columnwidth]{figures/results/MI_1440,1,4_2019-12-08_1.png}
  \caption{08-12-2019. Sampling period : $d=1$ minute. Lead-lag graph for $T = 2d$.}
  \label{fig:lag_net_1_2_08-12-2019}
\end{minipage}
\centering
\begin{minipage}{.60\columnwidth}
  \centering
  \includegraphics[width=.8\columnwidth]{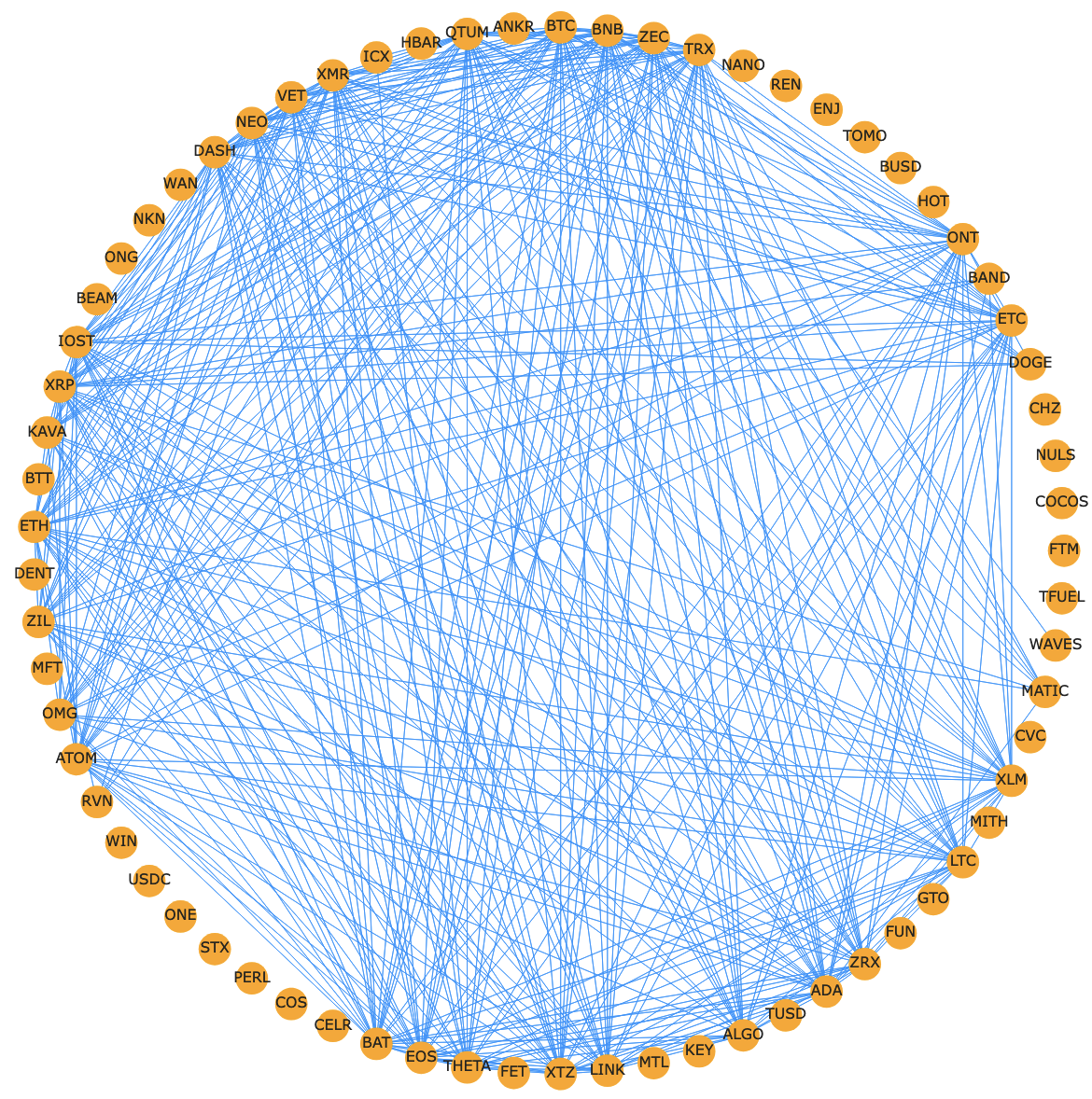}
  \caption{2020-08-08. Sampling period : $d=1$ minute. Lead-lag graph for $T = 0d$.}
  \label{fig:lag_net_1_0_2020-08-08}
\end{minipage}%
\hspace{0.1cm}
\begin{minipage}{.60\columnwidth}
  \centering
  \includegraphics[width=.8\columnwidth]{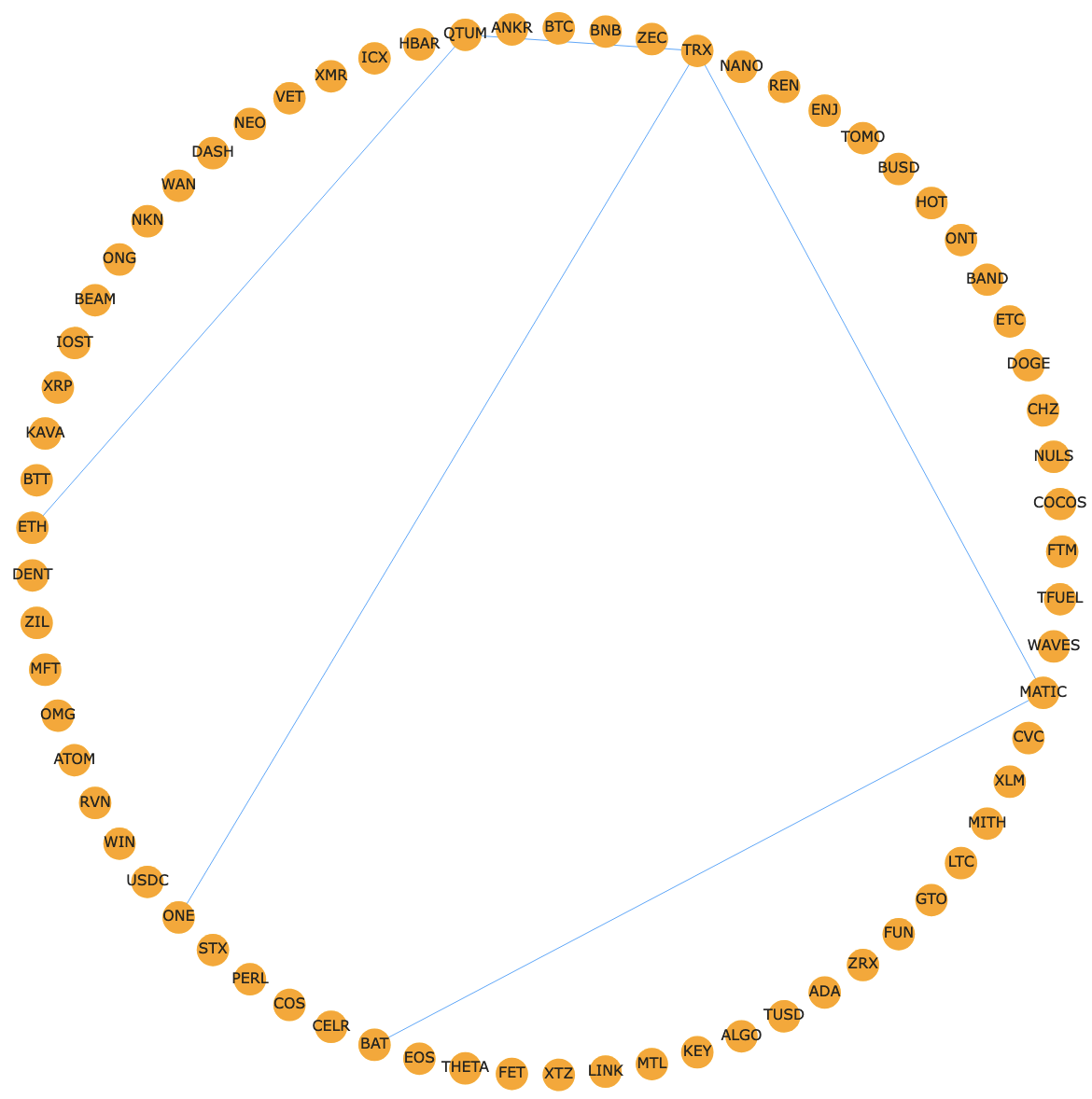}
  \caption{2020-08-08. Sampling period : $d=1$ minute. Lead-lag graph for $T = 1d$.}
  \label{fig:lag_net_1_1_2020-08-08}
\end{minipage}%
\hspace{0.1cm}
\begin{minipage}{.60\columnwidth}
  \centering
  \includegraphics[width=.8\columnwidth]{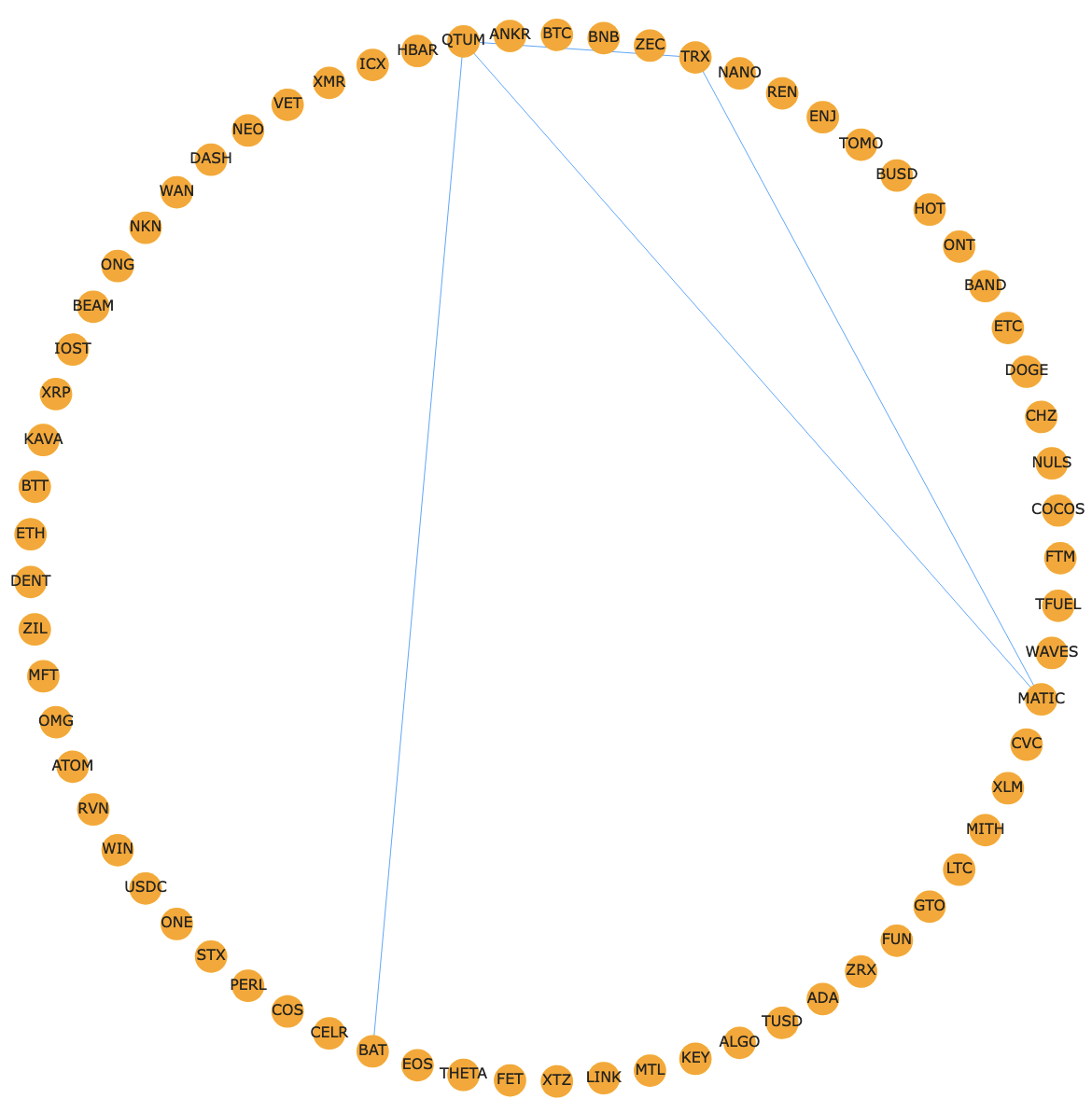}
  \caption{2020-08-08. Sampling period : $d=1$ minute. Lead-lag graph for $T = 2d$.}
  \label{fig:lag_net_1_2_2020-08-08}
\end{minipage}
\centering
\begin{minipage}{.60\columnwidth}
  \centering
  \includegraphics[width=.8\columnwidth]{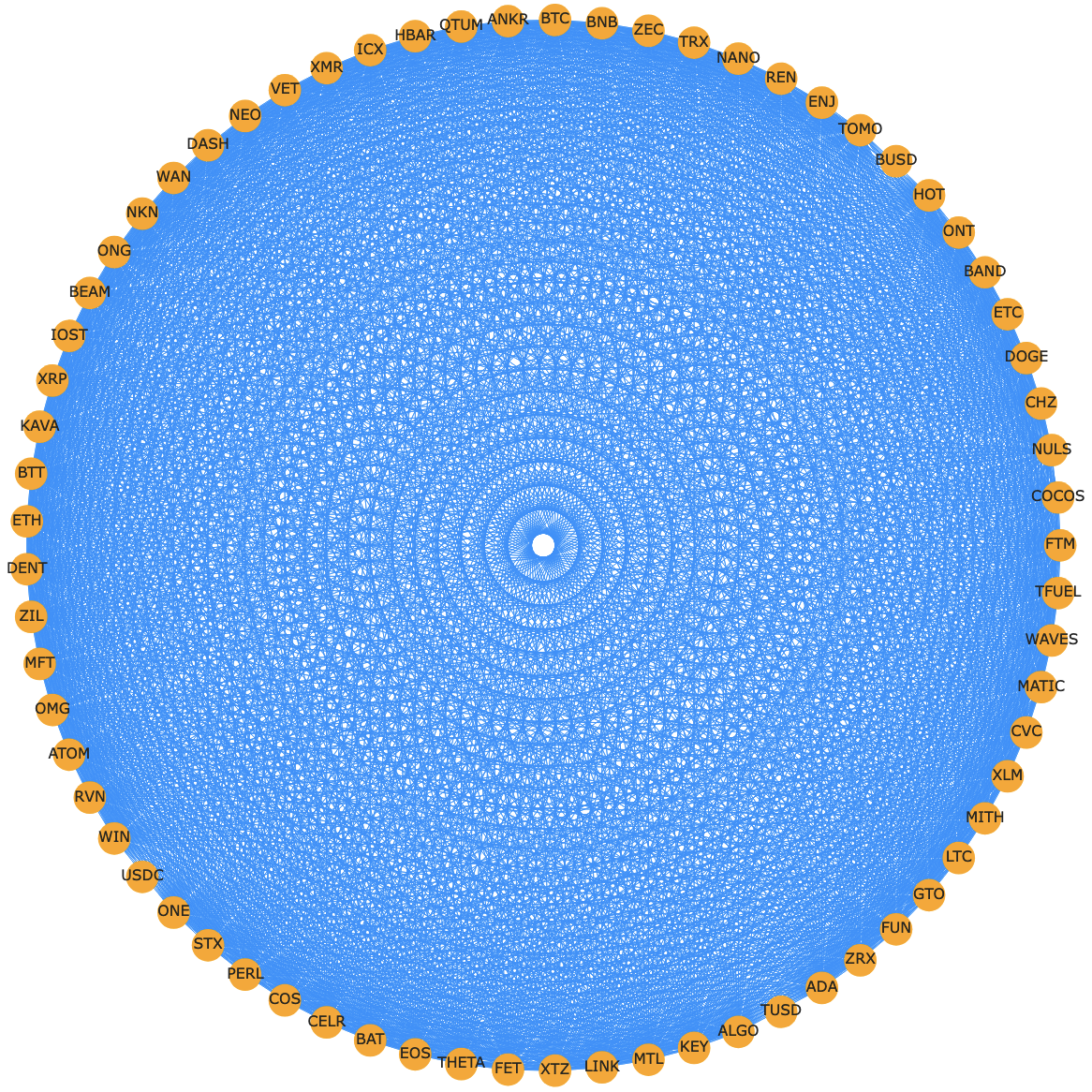}
  \caption{2021-09-08. Sampling period : $d=1$ minute. Lead-lag graph for $T = 0d$.}
  \label{fig:lag_net_1_0_2021-09-08}
\end{minipage}%
\hspace{0.1cm}
\begin{minipage}{.60\columnwidth}
  \centering
  \includegraphics[width=.8\columnwidth]{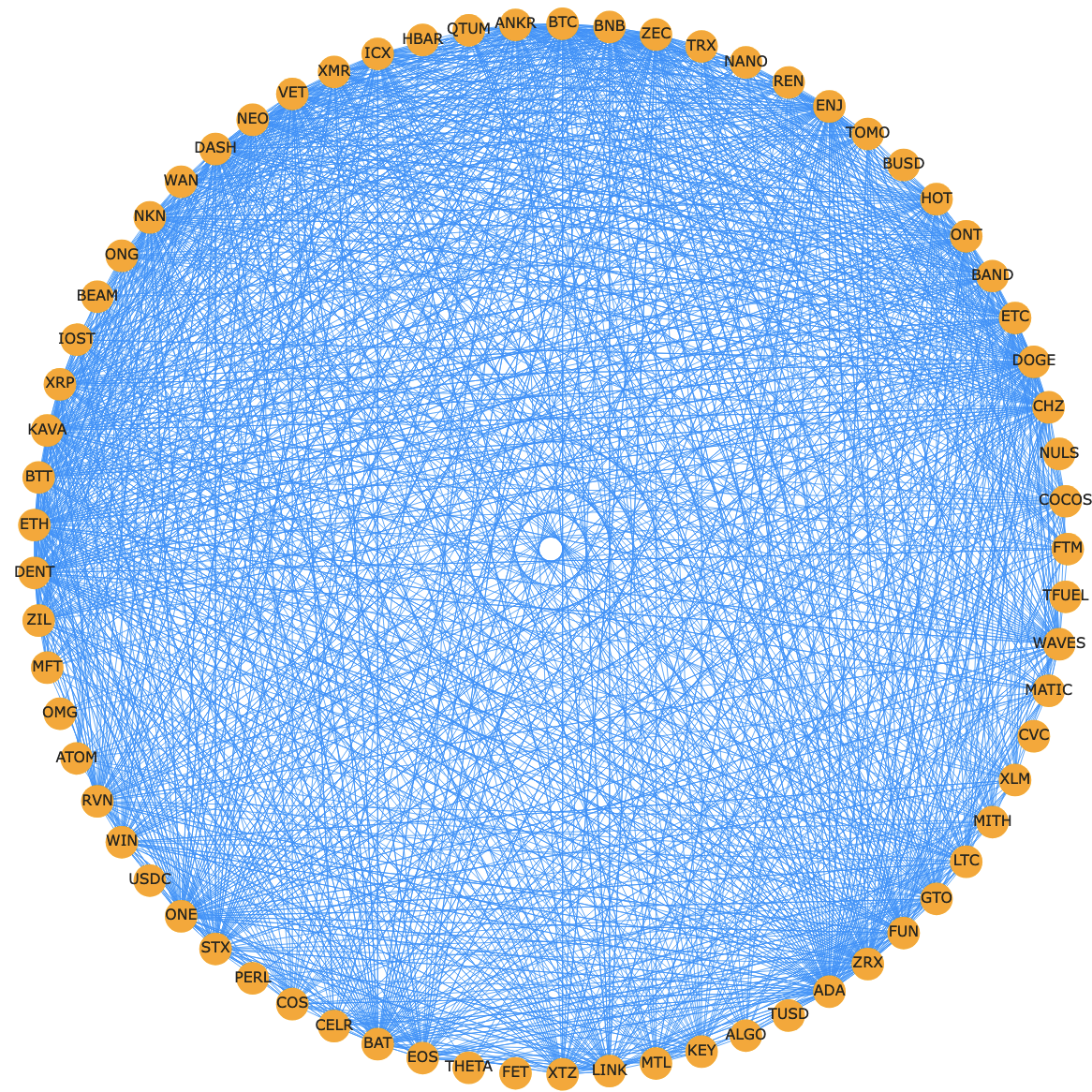}
  \caption{2021-09-08. Sampling period : $d=1$ minute. Lead-lag graph for $T = 1d$.}
  \label{fig:lag_net_1_1_2021-09-08}
\end{minipage}%
\hspace{0.1cm}
\begin{minipage}{.60\columnwidth}
  \centering
  \includegraphics[width=.8\columnwidth]{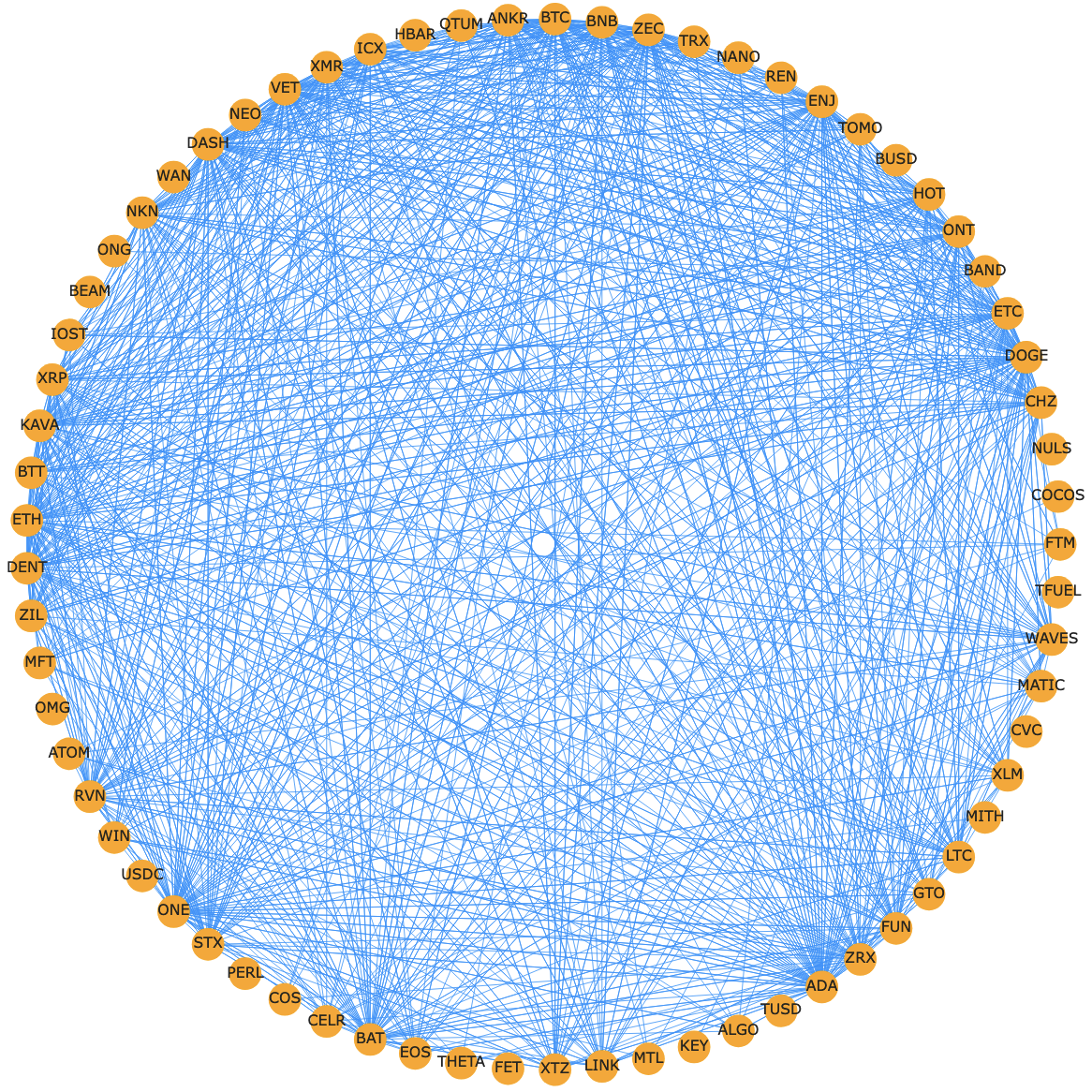}
  \caption{2021-09-08. Sampling period : $d=1$ minute. Lead-lag graph for $T = 2d$.}
  \label{fig:lag_net_1_2_2021-09-08}
\end{minipage}
\end{figure*}

\begin{figure*}[]
\centering
\begin{minipage}{.60\columnwidth}
  \centering
  \includegraphics[width=.8\columnwidth]{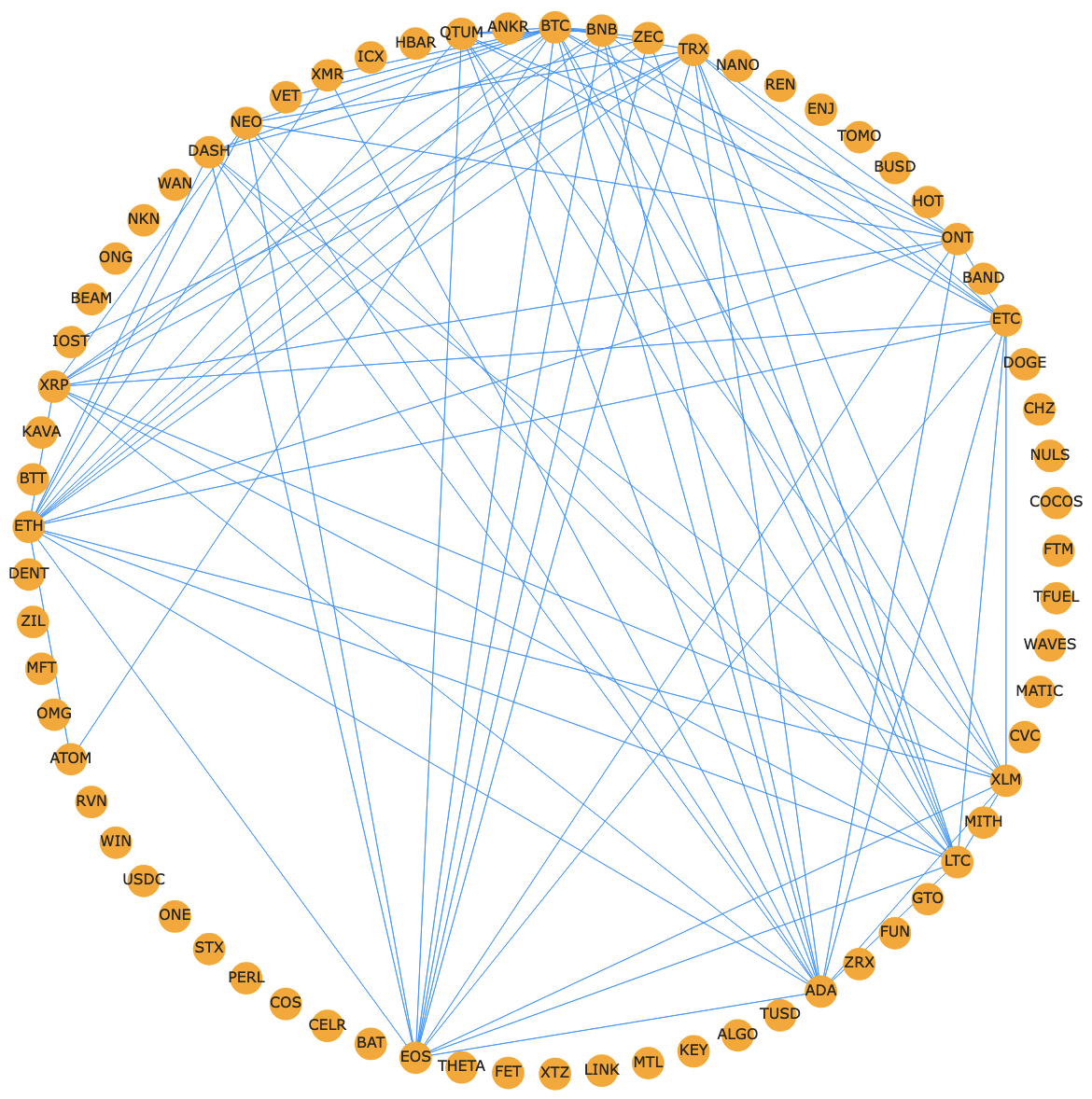}
  \caption{08-12-2019. Sampling period : $d=5$ minute. Lead-lag graph for $T = 0d$.}
  \label{fig:lag_net_5_0_08-12-2019}
\end{minipage}%
\hspace{0.1cm}
\begin{minipage}{.60\columnwidth}
  \centering
  \includegraphics[width=.8\columnwidth]{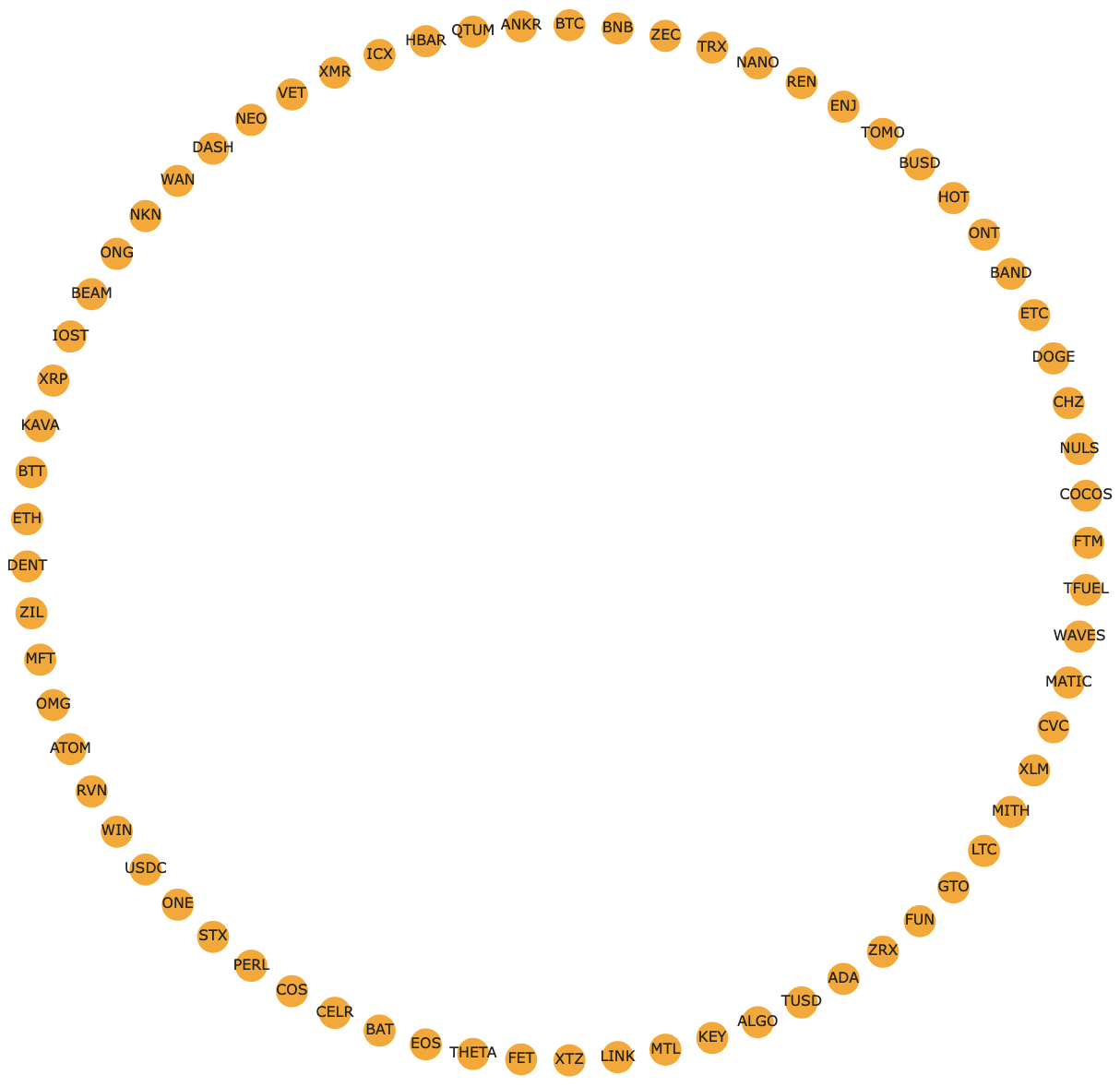}
  \caption{08-12-2019. Sampling period : $d=5$ minute. Lead-lag graph for $T = 1d$.}
  \label{fig:lag_net_5_1_08-12-2019}
\end{minipage}%
\hspace{0.1cm}
\begin{minipage}{.60\columnwidth}
  \centering
  \includegraphics[width=.8\columnwidth]{figures/results/MI_500,5,4_2019-12-08_1.png}
  \caption{08-12-2019. Sampling period : $d=5$ minute. Lead-lag graph for $T = 2d$.}
  \label{fig:lag_net_5_2_08-12-2019}
\end{minipage}
\centering
\begin{minipage}{.60\columnwidth}
  \centering
  \includegraphics[width=.8\columnwidth]{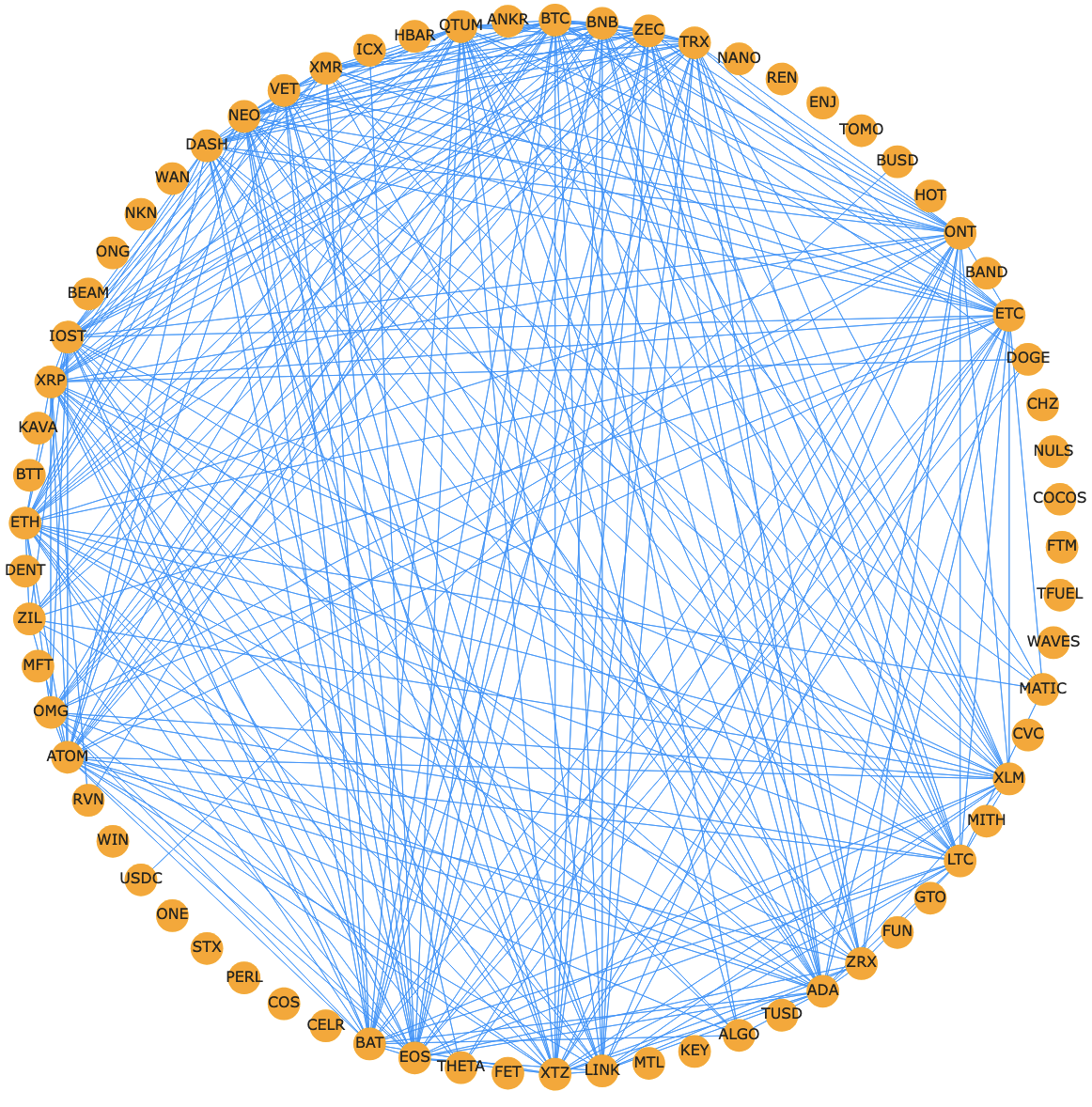}
  \caption{2020-08-08. Sampling period : $d=5$ minute. Lead-lag graph for $T = 0d$.}
  \label{fig:lag_net_5_0_2020-08-08}
\end{minipage}%
\begin{minipage}{.60\columnwidth}
  \centering
  \includegraphics[width=.8\columnwidth]{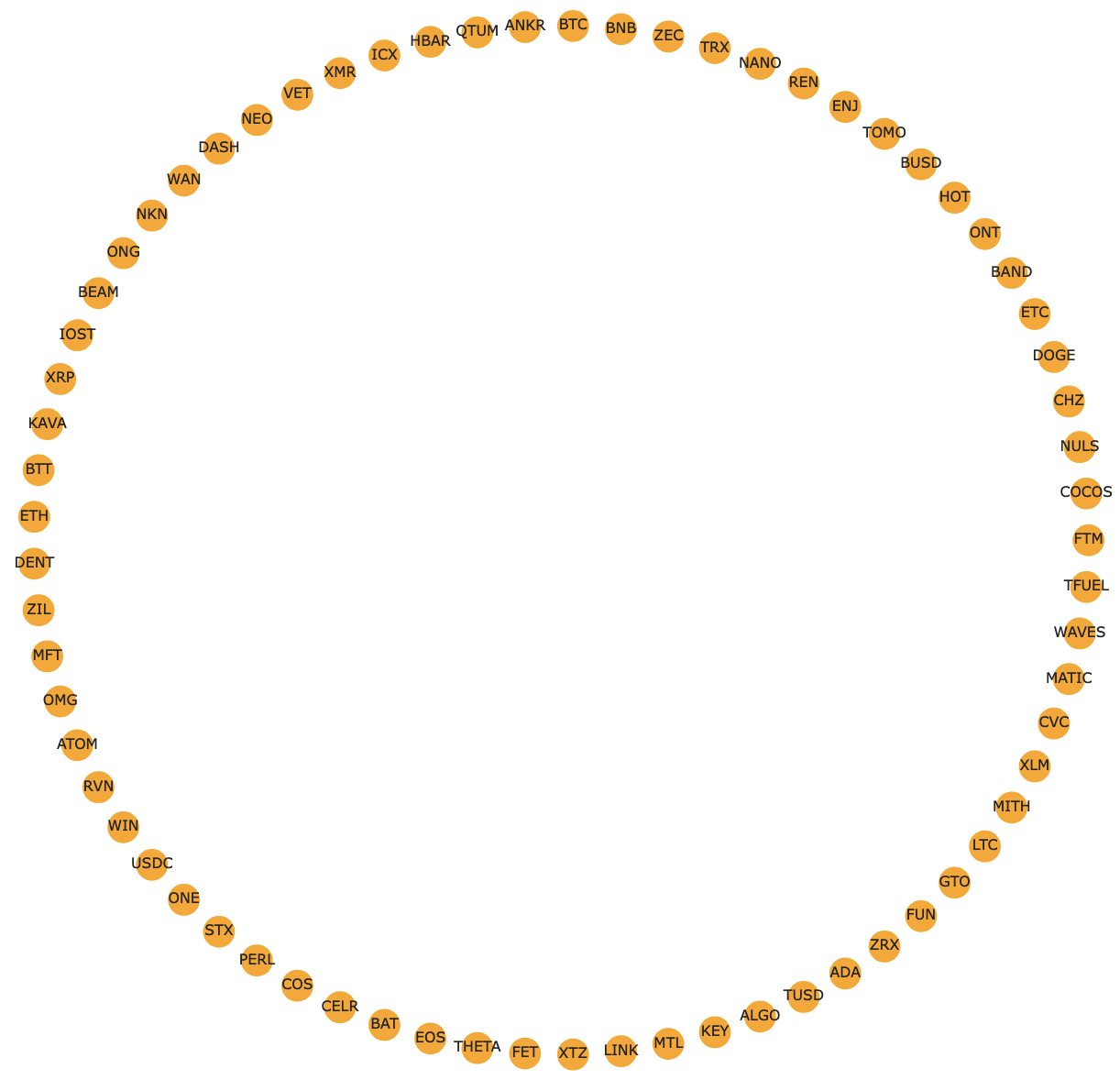}
  \caption{2020-08-08. Sampling period : $d=5$ minute. Lead-lag graph for $T = 1d$.}
  \label{fig:lag_net_5_1_2020-08-08}
\end{minipage}%
\hspace{0.1cm}
\begin{minipage}{.60\columnwidth}
  \centering
  \includegraphics[width=.8\columnwidth]{figures/results/MI_500,5,4_2020-08-08_1.png}
  \caption{2020-08-08. Sampling period : $d=5$ minute. Lead-lag graph for $T = 2d$.}
  \label{fig:lag_net_5_2_2020-08-08}
\end{minipage}
\centering
\begin{minipage}{.60\columnwidth}
  \centering
  \includegraphics[width=.8\columnwidth]{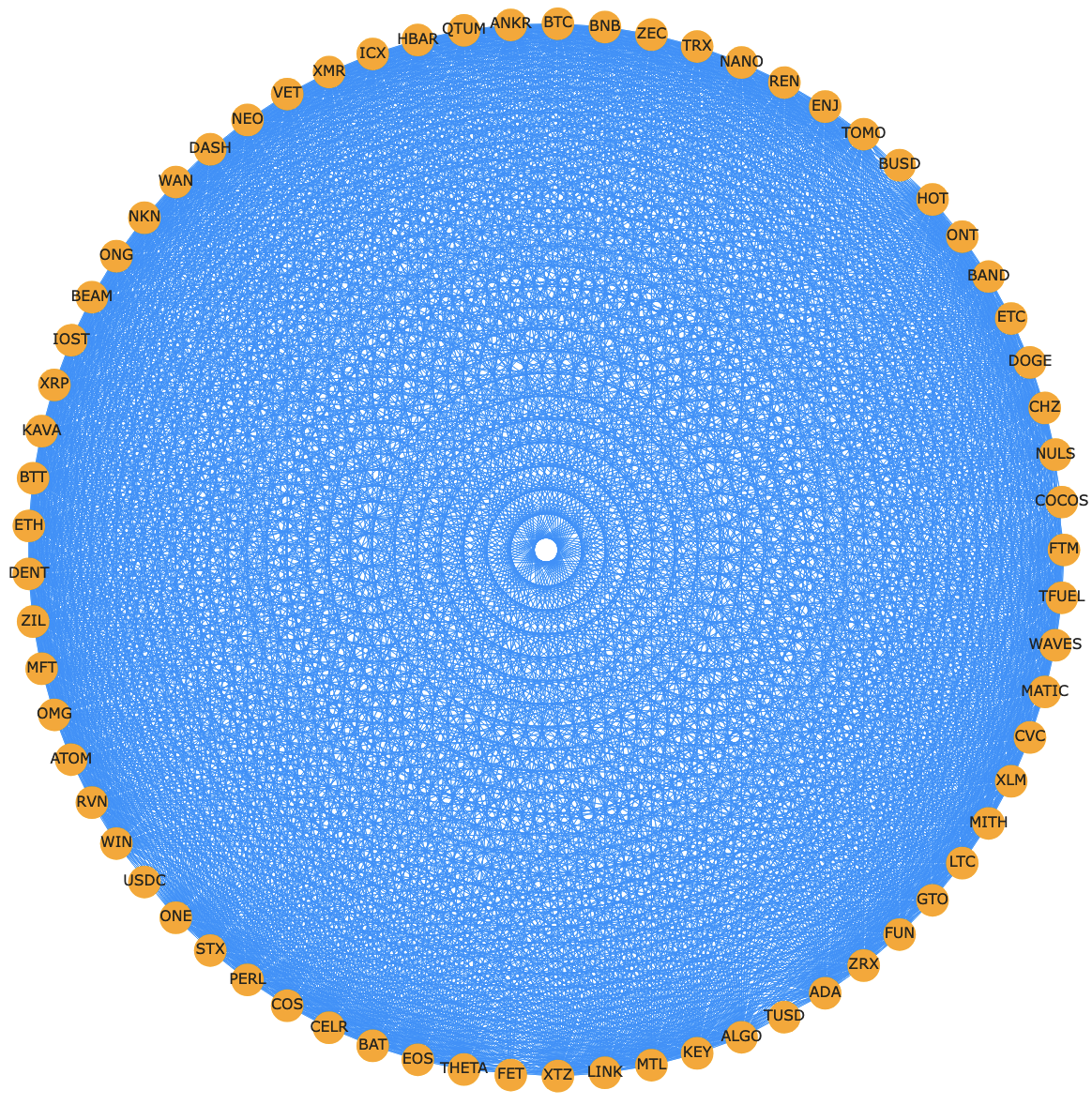}
  \caption{2021-09-08. Sampling period : $d=5$ minute. Lead-lag graph for $T = 0d$.}
  \label{fig:lag_net_5_0_2021-09-08}
\end{minipage}%
\hspace{0.1cm}
\begin{minipage}{.60\columnwidth}
  \centering
  \includegraphics[width=.8\columnwidth]{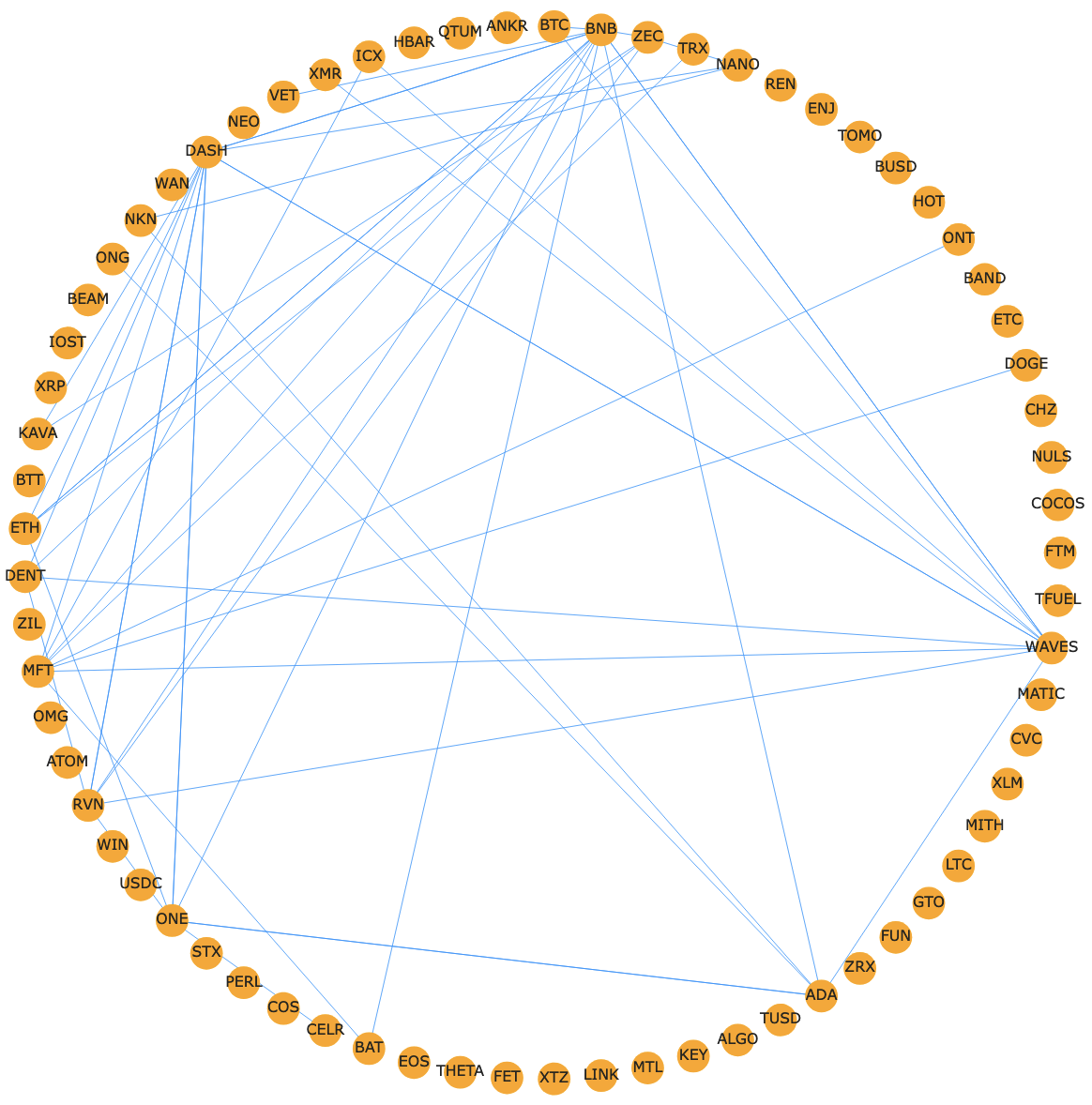}
  \caption{2021-09-08. Sampling period : $d=5$ minute. Lead-lag graph for $T = 1d$.}
  \label{fig:lag_net_5_1_2021-09-08}
\end{minipage}%
\hspace{0.1cm}
\begin{minipage}{.60\columnwidth}
  \centering
  \includegraphics[width=.8\columnwidth]{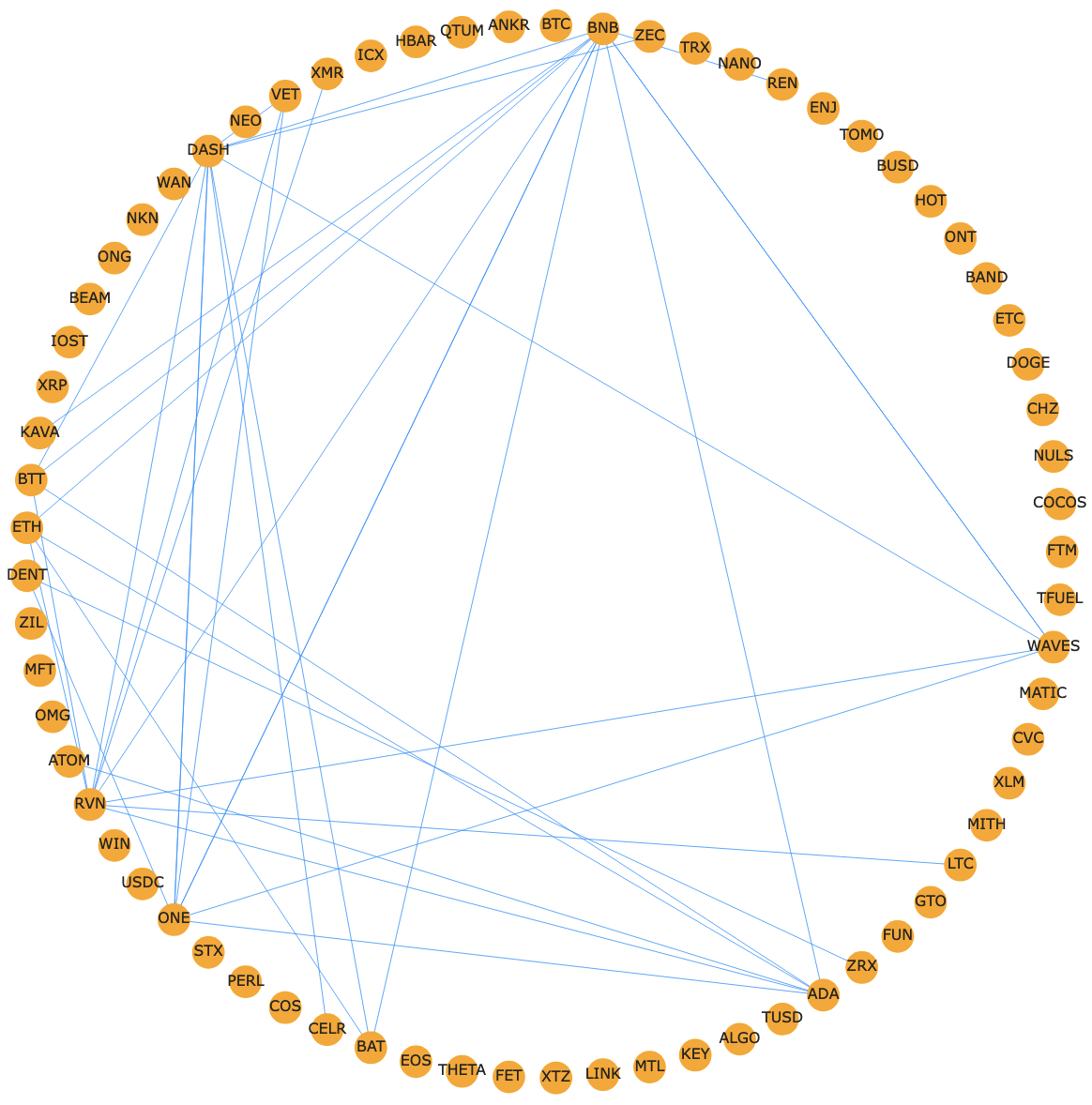}
  \caption{2021-09-08. Sampling period : $d=5$ minute. Lead-lag graph for $T = 2d$.}
  \label{fig:lag_net_5_2_2021-09-08}
\end{minipage}
\end{figure*}